%% file: main.tex
\PassOptionsToPackage{table}{xcolor}
\documentclass{article}
\usepackage[T1]{fontenc}
\usepackage{iclr2027_conference,times}
\input{math_commands.tex}

\usepackage{xcolor}
\usepackage[hidelinks]{hyperref}
\usepackage{url}
\usepackage{microtype}
\usepackage{amsmath,amssymb}
\usepackage{graphicx}
\usepackage{caption}
\usepackage{booktabs}
\usepackage{array}
\usepackage{multirow}
\usepackage{tabularx}
\usepackage[ruled,vlined]{algorithm2e}
\usepackage[most]{tcolorbox}
\usepackage{enumitem}
\usepackage{float}
\usepackage{listings}

\newcounter{proposition}

\usepackage{xcolor}
\definecolor{citeblue}{RGB}{0,0,180}

\hypersetup{
    colorlinks=true,
    allcolors=citeblue,
    breaklinks=true,
    pdfborder={0 0 0}
}

\definecolor{triton_green}{RGB}{4,106,56}
\definecolor{allocator_blue}{RGB}{50,72,130}
\definecolor{critic_purple}{RGB}{82,64,118}
\definecolor{smoke_brown}{RGB}{132,78,41}
\definecolor{schema_steel}{RGB}{68,91,112}
\definecolor{code_border}{RGB}{174,188,202}
\definecolor{code_bg}{RGB}{247,249,252}
\definecolor{code_keyword}{RGB}{196,53,42}
\definecolor{code_comment}{RGB}{42,132,78}
\definecolor{code_string}{RGB}{115,88,35}

\newcommand{\algmark}[1]{}
\SetKwComment{tcp}{}{}

\newcommand{\drawalgband}[2]{}
\newtcolorbox{experiencebox}[2][]{%
  enhanced,
  colback=triton_green!2,
  colframe=triton_green!28,
  boxrule=0.45pt,
  arc=1.5pt,
  left=5pt,
  right=5pt,
  top=2pt,
  bottom=3pt,
  boxsep=1pt,
  before skip=8pt,
  after skip=8pt,
  fontupper=\small,
  fonttitle=\small\bfseries,
  colbacktitle=triton_green!14,
  coltitle=black,
  titlerule=0pt,
  toptitle=2pt,
  bottomtitle=2pt,
  title={#2},
  #1
}

\lstdefinestyle{evocode}{%
  basicstyle=\ttfamily\footnotesize,
  breaklines=true,
  breakatwhitespace=false,
  columns=fullflexible,
  keepspaces=true,
  frame=single,
  rulecolor=\color{code_border},
  framerule=0.4pt,
  framesep=8pt,
  backgroundcolor=\color{code_bg},
  keywordstyle=\color{code_keyword}\bfseries,
  commentstyle=\color{code_comment}\itshape,
  stringstyle=\color{code_string},
  showstringspaces=false,
  captionpos=b,
  xleftmargin=0pt,
  framexleftmargin=6pt,
  framexrightmargin=6pt,
  framextopmargin=5pt,
  framexbottommargin=5pt,
  aboveskip=6pt,
  belowskip=6pt
}
\tcbset{
  evocritic listing base/.style={
    enhanced,
    breakable,
    listing only,
    boxrule=0.65pt,
    arc=2.5pt,
    outer arc=2.5pt,
    left=8pt,
    right=8pt,
    top=8pt,
    bottom=8pt,
    boxsep=0pt,
    before skip=9pt,
    after skip=9pt,
    fonttitle=\small\bfseries,
    coltitle=white,
    titlerule=0pt,
    toptitle=4pt,
    bottomtitle=4pt,
    listing options={
      basicstyle=\ttfamily\scriptsize,
      breaklines=true,
      breakatwhitespace=true,
      columns=fullflexible,
      keepspaces=true,
      showstringspaces=false,
      frame=none,
      backgroundcolor={},
      aboveskip=0pt,
      belowskip=0pt,
      numbers=none
    }
  }
}

\newtcblisting{promptlisting}[2][]{%
  evocritic listing base,
  colback=allocator_blue!3,
  colframe=allocator_blue!82!black,
  colbacktitle=allocator_blue!92!black,
  title={\strut #2},
  #1
}

\newtcblisting{criticpromptlisting}[2][]{%
  evocritic listing base,
  colback=critic_purple!3,
  colframe=critic_purple!82!black,
  colbacktitle=critic_purple!92!black,
  title={\strut #2},
  #1
}

\newtcblisting{smokepromptlisting}[2][]{%
  evocritic listing base,
  colback=smoke_brown!3,
  colframe=smoke_brown!82!black,
  colbacktitle=smoke_brown!92!black,
  title={\strut #2},
  #1
}

\newtcblisting{codebox}[1][]{%
  enhanced,
  breakable,
  listing only,
  center,
  width=0.94\linewidth,
  before skip=2pt,
  after skip=6pt,
  colback=code_bg,
  colframe=code_border,
  boxrule=0.45pt,
  arc=1.5pt,
  outer arc=1.5pt,
  left=5pt,
  right=4pt,
  top=4pt,
  bottom=4pt,
  boxsep=0pt,
  listing options={
    language=Python,
    basicstyle=\ttfamily\scriptsize,
    breaklines=true,
    breakatwhitespace=false,
    columns=fullflexible,
    keepspaces=true,
    showstringspaces=false,
    frame=none,
      backgroundcolor={},
      aboveskip=0pt,
      belowskip=0pt,
    numbers=none,
    keywordstyle=\color{code_keyword}\bfseries,
    commentstyle=\color{code_comment}\itshape,
    stringstyle=\color{code_string}
  },
  #1
}

\newtcblisting{casecardbox}[2][]{%
  enhanced,
  listing only,
  colback=schema_steel!2,
  colframe=schema_steel!32,
  colbacktitle=schema_steel!72!black,
  coltitle=white,
  fonttitle=\bfseries\footnotesize,
  boxrule=0.45pt,
  arc=1.5pt,
  outer arc=1.5pt,
  left=5pt,
  right=4pt,
  top=4pt,
  bottom=4pt,
  boxsep=0pt,
  before skip=3pt,
  after skip=3pt,
  titlerule=0pt,
  toptitle=3pt,
  bottomtitle=3pt,
  title={\strut #2},
  listing options={
    basicstyle=\ttfamily\scriptsize,
    breaklines=true,
    breakatwhitespace=false,
    columns=fullflexible,
    keepspaces=true,
    showstringspaces=false,
    frame=none,
      backgroundcolor={},
      aboveskip=0pt,
      belowskip=0pt,
    numbers=none
  },
  #1
}

\newtcblisting{schemalisting}[2][]{%
  enhanced,
  breakable,
  listing only,
  colback=schema_steel!3,
  colframe=schema_steel!45,
  colbacktitle=schema_steel!78!black,
  coltitle=white,
  fonttitle=\small\bfseries,
  boxrule=0.55pt,
  arc=2pt,
  outer arc=2pt,
  left=8pt,
  right=8pt,
  top=7pt,
  bottom=7pt,
  boxsep=0pt,
  before skip=8pt,
  after skip=8pt,
  titlerule=0pt,
  toptitle=4pt,
  bottomtitle=4pt,
  title={\strut #2},
  listing options={
    escapeinside={(*@}{@*)},
    basicstyle=\ttfamily\scriptsize,
    breaklines=true,
    breakatwhitespace=false,
    columns=fullflexible,
    keepspaces=true,
    showstringspaces=false,
    frame=none,
      backgroundcolor={},
      aboveskip=0pt,
      belowskip=0pt,
    numbers=none
  },
  #1
}

\newcommand{\experiencefield}[3]{%
  \noindent\textcolor{#1}{\textbf{\scriptsize #2}}\par\nobreak\vspace{0.5pt}%
  \noindent#3\par\vspace{2pt}%
}

\title{EvoAlloc: A Self-Evolving Resource Allocation Agent for Efficient Program Evolution}

\author{%
    \makebox[\dimexpr\textwidth-2\tabcolsep\relax][s]{%
        Yanning Dai\textsuperscript{1}\hfill
        Yuhui Wang\textsuperscript{1}\thanks{Corresponding author.}\hfill 
        Nanbo Li\textsuperscript{1}\hfill
        Wenyi Wang\textsuperscript{1,2}\hfill
        J\"urgen Schmidhuber\textsuperscript{1}%
    }\\[5pt]
    {%
    \normalfont
    \begin{tabular}{@{}l@{}}
        ${}^{1}$~Center of Excellence for Generative AI, King Abdullah University of Science and \\
        \hspace{.5em} Technology (KAUST), Thuwal, Saudi Arabia.\\
        $^2$~Sakana AI, Tokyo, Japan.
    \end{tabular}%
    }\vspace{.5em}
}

\newif\ifshowtmp
\showtmptrue

\ifshowtmp
    \newcommand{\yanning}[1]{\textcolor{blue}{(\textbf{Yanning}: #1)}}
    \newcommand{\yuhui}[1]{\textcolor{teal}{(\textbf{Yuhui}: #1)}}
    \newcommand{\juergen}[1]{\textcolor{blue}{(\textbf{Juergen}: #1)}}
\else
    \newcommand{\yanning}[1]{}
    \newcommand{\yuhui}[1]{}    
    \newcommand{\juergen}[1]{}    
\fi

\iclrfinalcopy

\begin{document}

\maketitle
\lhead{}
\suppressfloats[t]
\raggedbottom
\vspace{-12pt}
\begin{abstract}
LLM-based program evolution relies on evaluation feedback to guide the iterative search for high-performing programs. However, evaluation is often computationally expensive, making it essential to allocate limited resources to candidates that can most effectively advance the search. Existing LLM-based methods typically rely on fixed allocation strategies throughout the search, potentially wasting resources on low-value candidates while overlooking promising ones. We propose \textbf{EvoAlloc}, a self-evolving resource-allocation agent that learns from search experience to revise its strategy for allocating computational resources across candidates. EvoAlloc periodically consolidates prior search and allocation outcomes into reusable experience, which informs subsequent strategy revisions. It further uses a counterfactual exploration mechanism to occasionally evaluate candidates denied resources by the allocator, revealing their outcomes to enrich its experience for future strategy updates. Across coding and agent-harness optimization benchmarks, EvoAlloc requires $59$--$82$\% fewer full evaluations and $61$--$89$\% fewer total LLM tokens to reach baseline-level performance. Moreover, under the same full-evaluation budget, EvoAlloc achieves $8.7$--$12.0$\% higher final performance.
\end{abstract}

\vspace{-10pt}
\begin{figure}[h!]
    \centering
    \vspace{10pt}
    \captionsetup{skip=2pt}
    \includegraphics[width=0.98\linewidth]{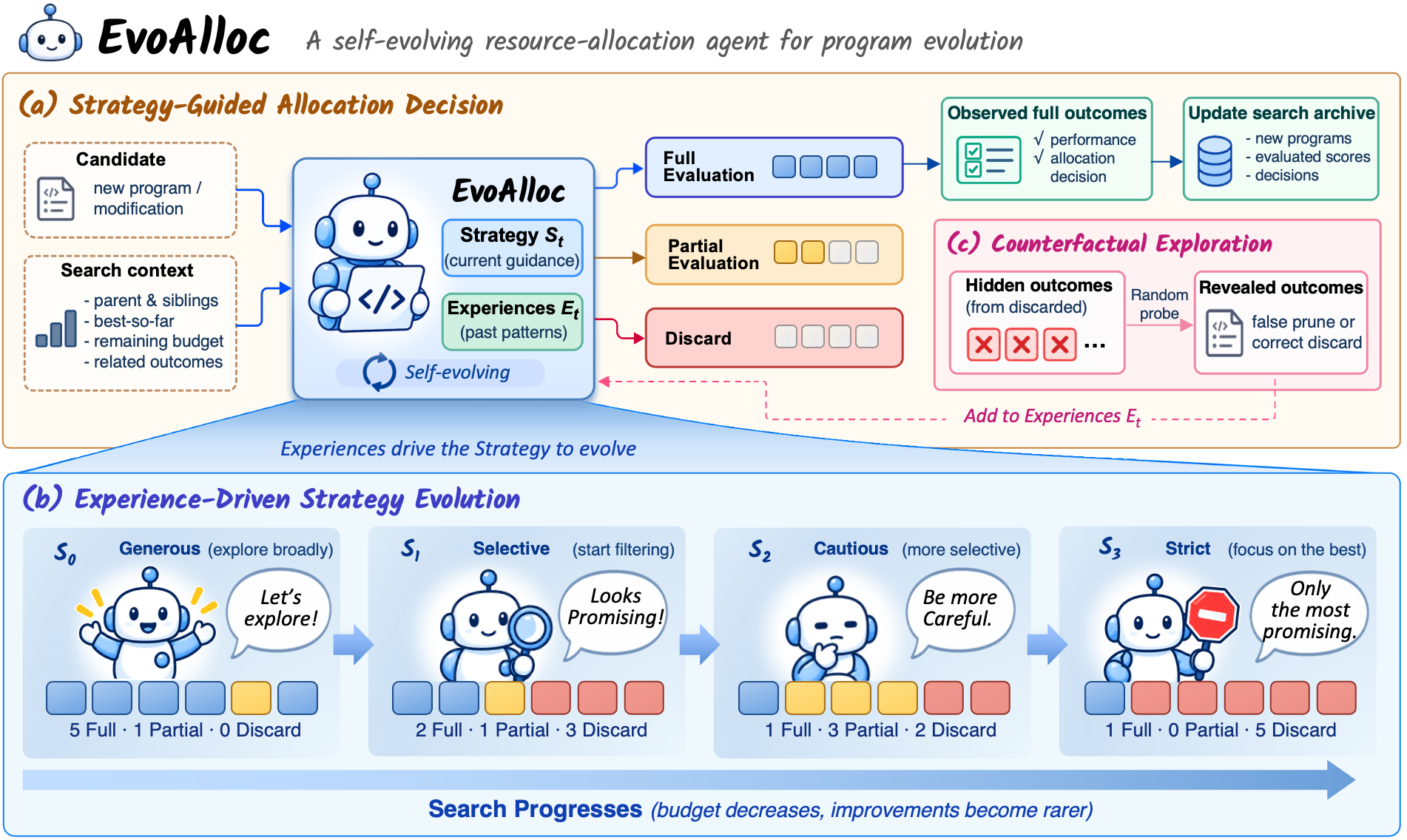}
    \vspace{2pt}
    \caption{\textbf{EvoAlloc, a self-evolving resource allocation agent for program evolution.} (a) Strategy-guided allocation decision, (b) Experience-driven Strategy evolution, and (c) counterfactual exploration jointly adapt how evaluation is allocated as search progresses.}
    \label{fig:method_overview}
\end{figure}

\section{Introduction}

Recent advances in LLM-based program evolution have established a general paradigm for discovering and refining executable programs through iterative cycles of generation, evaluation, and selection. By pairing the ability of language models to propose semantically meaningful code modifications with feedback from program execution, these methods have demonstrated strong potential across mathematical and algorithmic discovery \citep{funsearch}, scientific and systems optimization \citep{novikov2025alphaevolve}, and the automated improvement of coding agents \citep{lange2025shinkaevolve,dgm,hgm}. Evaluation plays a central role in this process because it directly shapes the trajectory of search. Evaluated programs populate the archive from which future parents are drawn \citep{wan2025loongflow}, whereas a promising candidate that goes unevaluated cannot influence what is explored next. At the same time, fully evaluating a candidate can be expensive, requiring execution over many benchmark instances or random seeds, calls to external models or solvers, or multi-step rollouts of LLM agents \citep{sia}. Deciding which candidates to evaluate, and how much evaluation each should receive, is therefore not merely a matter of saving compute but a central decision of the search itself.

Existing approaches largely make this decision using an allocation rule specified before search begins. The rule may be a predefined multi-fidelity procedure that promotes candidates through increasing evaluation budgets \citep{li2018hyperband,li2020asha}, or an LLM that judges, ranks, or filters candidates before full evaluation \citep{foster2026rpm}. While these methods may condition individual decisions on intermediate results or search history, the allocation strategy itself remains fixed. This leaves valuable information unused. Each evaluated candidate reveals whether the decision to evaluate it was worthwhile, and as search proceeds, these outcomes accumulate into experience about which allocation choices advanced the search. Such experience matters because the value of evaluating a candidate is not a property of the candidate alone. It depends on the evolving state of the search. As the incumbent improves, useful improvements become rarer and the remaining budget shrinks, so what is worth evaluating changes over time. We argue that the allocator should therefore co-evolve with the search, improving its strategy from the outcomes of its own decisions.

Building on this view, we introduce EvoAlloc, a self-evolving allocation agent embedded in the program-evolution loop (Figure~\ref{fig:method_overview}). EvoAlloc maintains an explicit allocation strategy that determines whether and how much to evaluate each candidate. Rather than keeping this strategy fixed, EvoAlloc evolves it as the search proceeds, drawing on a reflective memory that distills the search history and past allocation outcomes into reusable experience. Proposed revisions are adopted when shared validation favors them,
prioritizing fewer missed new-best candidates and then lower evaluation cost. This self-evolution loop, however, creates a selective-feedback challenge: the agent's own decisions determine which outcomes become observable. EvoAlloc therefore incorporates counterfactual exploration to gather additional experience beyond what its current strategy would otherwise generate. Across coding and agent-harness optimization benchmarks, EvoAlloc improves final performance by $8.7$--$12.0$\% over baseline under the same evaluation budget, while requiring $59$--$82$\% fewer full evaluations and $61$--$89$\% fewer recorded LLM tokens to match the baseline's final performance. Ablations confirm the contributions of strategy evolution and counterfactual exploration.

Our main contributions are:

\begin{enumerate}[leftmargin=1.45em,itemsep=2pt,parsep=0pt,topsep=2pt,partopsep=0pt]
    \item \textbf{We introduce a view of evaluation allocation as an evolving component of program evolution.} We reframe allocation from fixed per-proposal judgments into an evolving decision process that uses the outcomes of prior allocation decisions to inform future ones.
    \item \textbf{We introduce EvoAlloc, a self-evolving resource-allocation agent.} EvoAlloc acts through an explicit strategy, accumulates reusable experience from its interaction with the search, explores counterfactual outcomes, and revises future allocation behavior.
    \item \textbf{We provide an empirical analysis of how evolving evaluation allocation reshapes program search.} Across coding and agent-harness benchmarks, we show that adaptive allocation improves search under limited budgets and analyze how it changes resource use, candidate coverage, and the resulting search trajectories.
\end{enumerate}

\input{sections/related_work}

\section{Problem Formulation}
\label{sec:formulation}
\label{sec:problem}

\subsection{Evaluation Allocation in Program Evolution}
\label{sec:formulation:allocation-agent}
\label{sec:stateful_allocation}

Consider a program-evolution process that iteratively generates candidate programs,
evaluates them, and uses the evaluated programs to guide subsequent generation.
We formulate evaluation allocation within this process as a sequential decision
problem solved by an allocation agent $\mathcal{A}$, which distributes a limited
evaluation budget across candidates to maximize the best program score reached by
the search.
At each search step $t$, a candidate program $x_t$ is generated, and the agent
observes a decision-time context $c_t$ containing information available before
evaluating $x_t$, such as the parent program and its score, the best score found
so far, and the remaining budget.

For candidate $x_t$, allocation may involve multiple decisions.
Let $j = 0, 1, \ldots, K_t$ index the decisions within its allocation episode,
and let $a_{t,j}$ denote the allocation action taken at decision $j$.
Each intermediate action $a_{t,j}$ ($j < K_t$) acquires evaluation evidence
$e_{t,j}$ about $x_t$, which becomes available to subsequent decisions, and the
final action $a_{t,K_t}$ either commits the candidate to full evaluation or stops
further evaluation.
The agent may condition each decision on the candidate, the decision-time context,
and the evidence acquired earlier in the same episode:
\begin{equation}
a_{t,j} = \mathcal{A}(x_t, c_t, e_{t,<j}),
\label{eq:allocation-action}
\end{equation}
where $e_{t,<j} := (e_{t,0}, \ldots, e_{t,j-1})$, with $e_{t,<0} := \varnothing$.
The resulting allocation episode is
$\tau_t = (a_{t,0}, e_{t,0}, \ldots, a_{t,K_t-1}, e_{t,K_t-1}, a_{t,K_t})$.
Let $y_t \in \mathbb{R}$ denote the full-evaluation score of $x_t$, which is
defined regardless of whether $x_t$ is actually fully evaluated.
The score observed from the allocation episode is
\begingroup
\setlength{\belowdisplayskip}{2pt}
\setlength{\belowdisplayshortskip}{2pt}
\[
y_t^{\mathrm{obs}}
=
\begin{cases}
y_t, & \text{if $x_t$ is fully evaluated},\\
\varnothing, & \text{otherwise}.
\end{cases}
\]
\endgroup
Only $y_t^{\mathrm{obs}}$ is available to the search and to the agent;
the scores of candidates that are not fully evaluated remain hidden.

Let $b_t$ denote the cumulative resource consumed after proposal $t$, measured,
for example, by completed full evaluations or cumulative LLM tokens.
Using this resource variable, we evaluate $\mathcal{A}$ through the resulting search
trajectory. Let $Q_0$ denote the best score among the initial programs. For any resource level
$b$, define
\begin{equation}
Q_{\mathcal{A}}^\star(b)
=
\max\left(
\{Q_0\}
\cup
\left\{
y_t^{\mathrm{obs}}
: t \in \mathbb{N},\;
b_t \leq b,\;
y_t^{\mathrm{obs}} \neq \varnothing
\right\}
\right),
\label{eq:best-score-resource}
\end{equation}
which gives the best fully evaluated program score obtained after consuming at
most resource $b$.
Given a total resource budget $B$, the objective is to choose an allocation
agent $\mathcal{A}$ that maximizes $\mathbb{E}[Q_{\mathcal{A}}^\star(B)]$,
where the expectation is over the stochasticity of both the search process and
the allocation agent. Conversely, for a target score $q$, the resource required
to reach it is
$C_q(\mathcal{A}) = \inf\{b \geq 0 : Q_{\mathcal{A}}^\star(b) \geq q\}$.
These objectives make evaluation allocation a search-level sequential decision problem rather than
independent proposal-level decisions. Resources saved on one candidate remain available for later
search, while fully evaluated candidates can influence which programs are explored next.

\subsection{Self-Evolving Allocation from Search Experience}
\label{sec:strategy-formulation}
\label{sec:formulation:strategy-evolution}
\label{sec:selective_feedback}

Across proposals, the agent accumulates the interaction history
$\mathcal{H}_t=\{(x_i,c_i,\tau_i,y_i^{\mathrm{obs}})\}_{i<t}$,
which contains all allocation interactions before search step $t$.
Here, $\tau_i$ records the allocation actions and intermediate evidence for
candidate $x_i$.
Episodes without Full evaluation are also included, with
$y_i^{\mathrm{obs}}=\varnothing$, although these candidates cannot participate
in subsequent program evolution.

We represent the allocation agent over the search as
$\mathcal{A}=(\mathcal{A}_t)_{t\ge0}$, where $\mathcal{A}_t$ denotes its
decision mechanism at search step $t$.
A fixed allocation agent may make different decisions as the proposal,
search context, and within-episode evidence change, while its decision
mechanism remains unchanged: $\mathcal{A}_t=\mathcal{A}_0$ for all $t$.
We call an allocation agent \emph{self-evolving} when it can also revise
this mechanism using the accumulated interaction history:
$\mathcal{A}_{t+1}=\mathcal{U}(\mathcal{A}_t,\mathcal{H}_{t+1})$,
where $\mathcal{U}$ denotes the agent-update process and leaves
$\mathcal{A}_t$ unchanged when no update is performed.
Section~\ref{sec:method} instantiates this formulation in EvoAlloc, where
$\mathcal{A}_t$ is guided by an explicit allocation Strategy $S_t$ and
reusable Experiences $E_t$ distilled from observed search outcomes.

\section{Method}
\label{sec:method}
\raggedbottom

\paragraph{EvoAlloc.}
EvoAlloc is a self-evolving evaluation-allocation agent for program evolution.
It follows an explicit strategy that allocates evaluation resources to each candidate
using the information available at decision time.
As outcomes accumulate, EvoAlloc distills prior allocation episodes into reusable
experience and uses it to evolve the strategy.
It further uses counterfactual exploration to reveal outcomes of selected candidates
that would otherwise remain unevaluated, providing additional feedback for subsequent
strategy evolution.
The following sections describe these three mechanisms: explicit evaluation allocation
(Section~\ref{sec:adaptive_routing}), experience-driven strategy evolution (Section~\ref{sec:allocation_state}), and counterfactual
exploration (Section~\ref{sec:counterfactual_correction}).

\begin{figure}[!t]
    \centering
    \includegraphics[width=\textwidth]{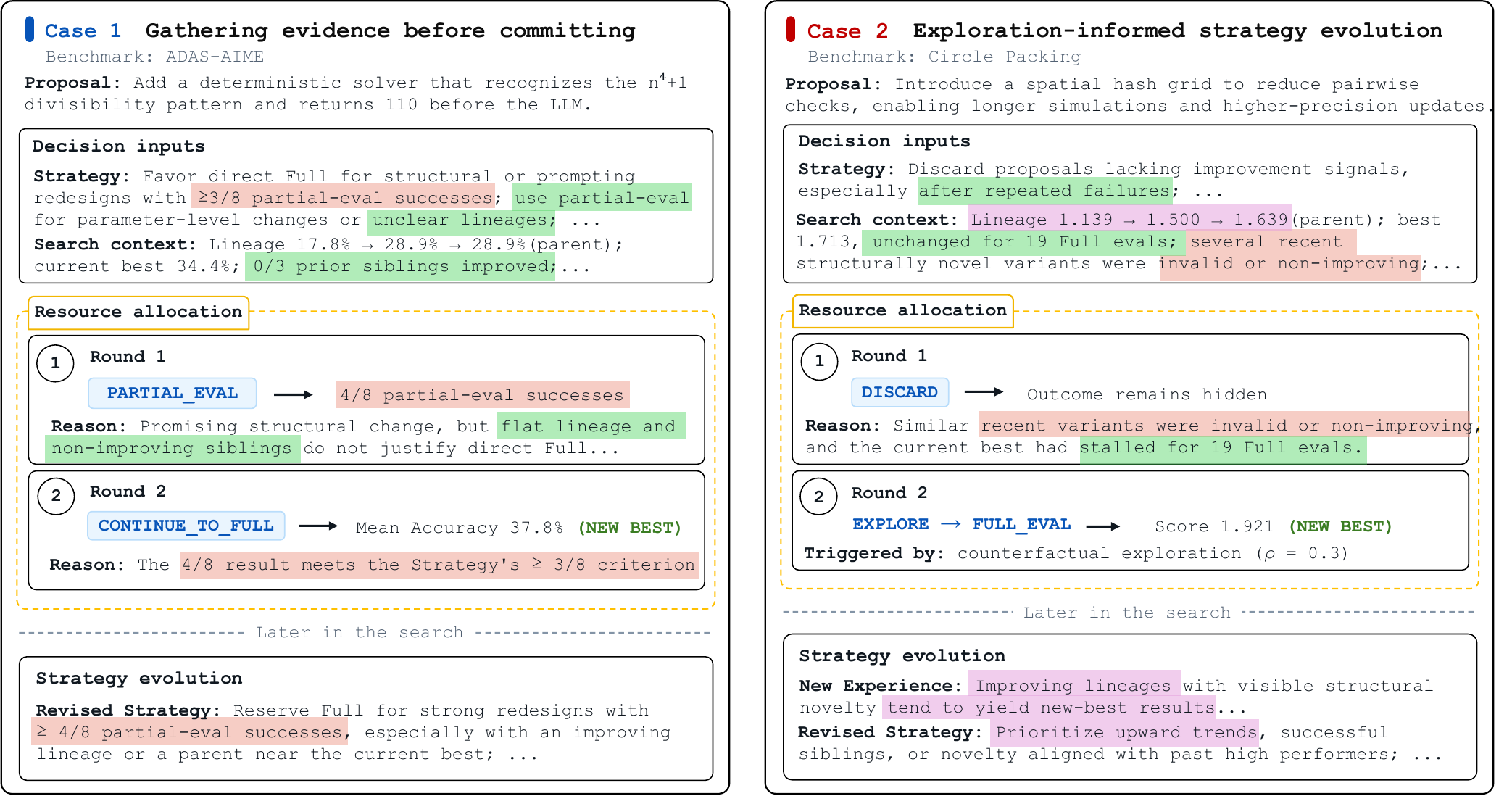}
    \caption{\textbf{Information flow between the strategy and allocation decisions in EvoAlloc.}
    Each case reads top to bottom: the strategy and search context inform the allocation
    agent's decisions, whose outcomes feed into later strategy updates.
    Proposals describe changes to the parent program.
    Matching highlights link related information across stages.
    Content is shortened for clarity. Strategy updates reflect feedback accumulated
    over multiple proposals, not the illustrated proposal alone.}
    \label{fig:case_studies}
\end{figure}

\subsection{Explicit Evaluation Allocation Strategy}
\label{sec:allocation-strategy}
\label{sec:allocation_strategy}
\label{sec:adaptive_routing}

EvoAlloc maintains an explicit natural-language allocation strategy $S_t$
that persists across candidates and is revised as the search progresses
(Sec.~\ref{sec:strategy-evolution}). Rather than making each allocation decision from its current
inputs alone, $S_t$ captures search-level guidance for allocating evaluation
resources. It specifies both the allocator's overall selectivity and
conditional guidance for choosing allocation actions under different search
contexts. Representative strategy trajectories are provided in Appendix~\ref{app:representative_guidance}.

In EvoAlloc, the agent $\mathcal{A}_t$ at search step $t$ is determined by its
current Strategy $S_t$ and Experiences $E_t$, which are provided to a fixed
allocator LLM together with the inputs of Eq.~(\ref{eq:allocation-action}).
Here, $c_t$ captures the decision-time search context, including the parent score, evaluated
sibling outcomes, current best score, and remaining budget, while $e_{t,<j}$ contains evidence
acquired earlier in the same allocation episode.
Across candidates, $E_t$ summarizes reusable evidence from observed outcomes, and $S_t$
provides the current run-level allocation guidance.
The allocation action space is task-dependent. In all settings, EvoAlloc may
choose \texttt{FULL\_EVAL} to commit the candidate to full evaluation or
\texttt{DISCARD} to stop further evaluation. When a meaningful lower-cost
intermediate evaluator is available, the agent may additionally choose
\texttt{PARTIAL\_EVAL} to acquire further evidence. After such evidence is
observed, the subsequent action is either \texttt{CONTINUE\_TO\_FULL} or
\texttt{STOP}. These actions instantiate the multi-step allocation episode
defined in Section~\ref{sec:formulation:allocation-agent}.

\subsection{Experience-Driven Strategy Evolution}
\label{sec:strategy-evolution}
\label{sec:allocation_state}

EvoAlloc implements the agent-update process $\mathcal{U}$ of Section~\ref{sec:formulation:strategy-evolution}
by updating two complementary components from observed outcomes.
Experiences $E_t$ summarize recurring patterns across observed cases, and the
Strategy $S_t$ translates these patterns into allocation criteria.

\paragraph{Experience updates.}
After every $k_E$ newly observed Full outcomes, the LLM consolidates the
new cases into $E_t$ by adding, updating, or removing Experiences.
Each Experience maintains the indices of its supporting and contradicting cases
and a status of \emph{tentative}, \emph{active}, or \emph{stale}.
The case indices are chronological, preserving how the evidence for an Experience
changes over time.
New patterns enter as \emph{tentative}; during later updates, the LLM may promote
well-supported Experiences to \emph{active} or mark outdated ones as
\emph{stale} as new evidence accumulates.
Stale Experiences remain as historical records but no longer guide allocation.

\paragraph{Strategy evolution.}
EvoAlloc starts from an initial strategy $S_0$ derived from a few example cases.
Each run begins with a short warm-up in which the allocator acts under $S_0$ but every
candidate is fully evaluated, seeding run-specific evidence.
Thereafter, every $k_S$ newly observed Full outcomes, EvoAlloc reflects on the current
strategy, Experiences, and recent cases to propose a challenger $\widetilde{S}$.
The challenger is validated against the current strategy on subsequent candidates that induce
different allocation paths under the two strategies (disagreements).
For each disagreement, EvoAlloc executes the union of both strategies' requested evaluations
and shares the resulting evidence.
Validation ends after two disagreements with an observed Full outcome.
Sharing evaluations prevents outcomes requested only by the challenger from being
censored in favor of the incumbent (Appendix~\ref{app:analysis}).
The two strategies are then compared lexicographically on the realized consequences of
their decisions: fewer missed new-best candidates, then fewer Full evaluations, then fewer
Partial evaluations.
The challenger is adopted if it is better on the first criterion where the two differ;
otherwise, the current strategy is retained.
This grounds strategy updates in observed allocation consequences rather than in LLM
judgment alone.
Implementation details and prompts are provided in Appendix~\ref{app:prompts}.

\subsection{Counterfactual Exploration}
\label{sec:counterfactual}
\label{sec:counterfactual-exploration}
\label{sec:counterfactual_correction}

Strategy evolution requires evaluation feedback, yet selective allocation itself determines which
feedback becomes observable. As defined in Sec.~\ref{sec:formulation:allocation-agent}, candidates that terminate without full evaluation
have $y_t^{\mathrm{obs}}=\varnothing$. The resulting feedback is therefore filtered by the current strategy:
systematic allocation errors can remain hidden because the corresponding candidates are never evaluated.
To expose such missing feedback, EvoAlloc introduces \emph{counterfactual exploration}. For each candidate
whose allocation episode terminates in \texttt{DISCARD} or \texttt{STOP}, EvoAlloc independently samples
$z_t \sim \mathrm{Bernoulli}(\rho)$,
where $\rho$ is the exploration rate ($0.3$ by default). When $z_t=1$, EvoAlloc executes
\texttt{EXPLORE\_EVAL}, fully evaluating the otherwise skipped candidate and revealing
$y_t^{\mathrm{obs}}=y_t$. Exploratory outcomes are marked separately from strategy-selected feedback.
Strong outcomes provide counterevidence to the current allocation criteria, while weak ones support them;
both inform subsequent experience and strategy updates. Once fully evaluated, an explored candidate
participates in the underlying program search as usual.

Together, these components form a closed loop: the Strategy guides the agent's decisions, and the outcomes of these decisions are used to revise the Strategy. Figure~\ref{fig:case_studies} illustrates this loop through two cases. In Case~1, the Strategy's evidence threshold leads the agent to request Partial evaluation before committing to Full evaluation, and accumulated outcomes later refine this threshold. In Case~2, counterfactual exploration reveals the outcome of a candidate the agent had discarded, exposing feedback that would otherwise remain unobserved and informing a revised Strategy.

\flushbottom
\section{Experiments}
\label{sec:experiments}

\subsection{Experimental Setup}
\label{sec:experimental_setup}

\paragraph{Implementation and protocol.}
\label{sec:evaluation_protocol}
We implement EvoAlloc within the ShinkaEvolve program-evolution framework
\citep{lange2025shinkaevolve},
with all methods sharing the search pipeline except for evaluation allocation.
Unless otherwise stated, experiments use five independent seeds.
We report mean best-so-far performance against Full evaluations and cumulative
LLM tokens, together with final best performance.
Our primary efficiency metric is matched-target cost. We first average best-so-far
scores and cumulative token usage across five seeds at each completed Full-evaluation
index. We then find the first index at which the mean score reaches a common target
and report that index and its corresponding mean cumulative token usage
(Appendix~\ref{app:resource_costs}).

\paragraph{Benchmarks.}
We evaluate EvoAlloc across program-evolution settings spanning
reasoning-harness optimization and solution-program optimization,
with differing access to intermediate evaluation feedback.
\textbf{ADAS-AIME} evolves programmatic reasoning harnesses around a fixed Qwen3-8B
solver on the 30 AIME 2024 problems.
Each candidate is scored by its mean accuracy over three independent runs.
A Full evaluation uses all 30 problems, while a Partial evaluation uses the first eight
before deciding whether to continue to Full.
\textbf{Circle Packing} evolves Python programs that arrange 26 non-overlapping circles
within the unit square to maximize the sum of their radii.
Its evaluator is deterministic, and since no lower-cost intermediate evaluation is available, the allocator chooses directly
between Full and Discard.

\paragraph{Baselines.}
We compare against four baselines within the same search framework.
\textbf{ShinkaEvolve}~\citep{lange2025shinkaevolve} fully evaluates every proposal that passes
the standard checks, providing a reference without evaluation allocation.
\textbf{One-Step LLM Allocator} uses the same allocator LLM and search context as
EvoAlloc, but makes each allocation decision directly without an explicit Strategy
or accumulated Experience.
\textbf{Lineage-UCB} is an LLM-free adaptive allocator that builds a Beta posterior
from parent-improvement outcomes in the local lineage and applies a fixed UCB rule
for evaluation allocation.
\textbf{RPM}~\citep{foster2026rpm} is a recent LLM-based evaluation-allocation method that
generates 15 proposals at each selection step and uses pairwise LLM comparisons in
a knockout tournament to select one for Full evaluation.
Implementation details, hyperparameters, token accounting, and baseline adaptations
are provided in Appendices~\ref{app:impl} and~\ref{app:detailed_results}.

\begin{figure}[!t]
    \centering
    \includegraphics[width=\textwidth]{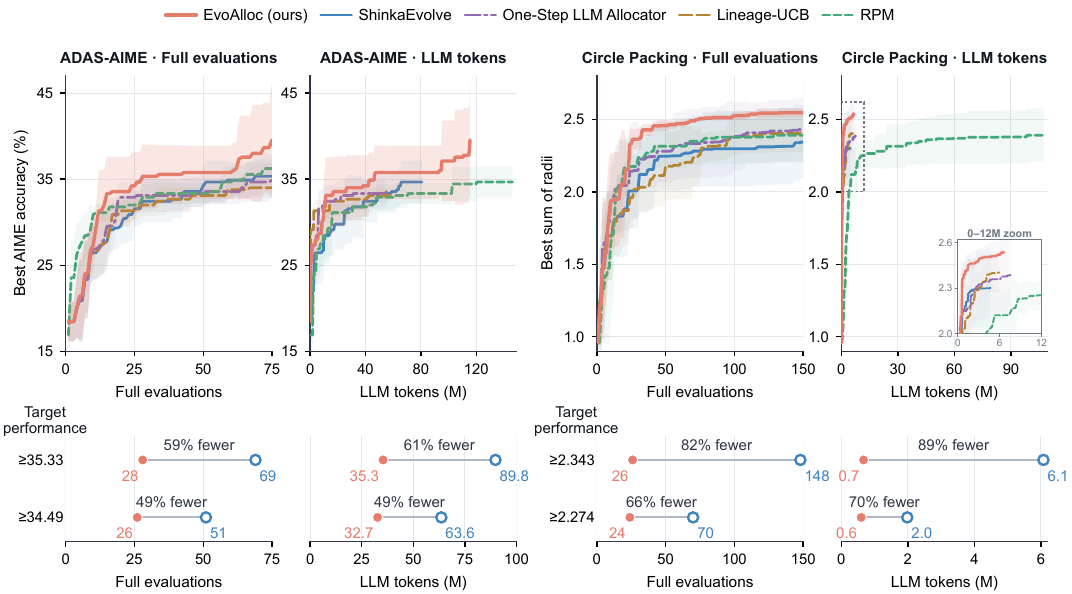}
    \caption{\textbf{Main results on ADAS-AIME and Circle Packing.}
    Curves show five-seed mean best-so-far performance ($\pm1$ std.) against Full evaluations
    and cumulative LLM tokens.
    Bottom panels report the resources required to reach ShinkaEvolve's final-performance
    and $95$\%-gain targets. Token costs are the mean cumulative usage at the first
    target-reaching Full-evaluation index of the evaluation-aligned mean performance curve.}
    \label{fig:main_sample_efficiency}
\end{figure}

\begin{figure}[!t]
\centering
\includegraphics[width=\textwidth]{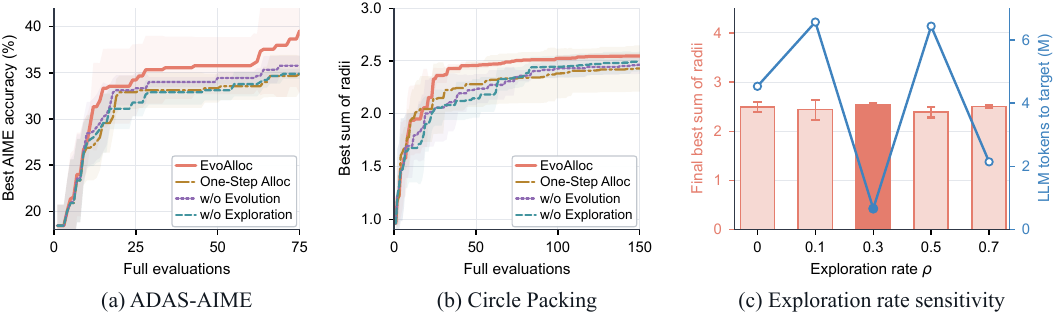}
\caption{\textbf{Component ablations and exploration-rate sensitivity.}
(a--b) Five-seed online ablations under matched Full-evaluation budgets; shading
denotes $\pm1$ standard deviation.
(c) Exploration-rate sweep on Circle Packing over $\rho \in \{0, 0.1, 0.3, 0.5, 0.7\}$; bars show the final best sum of radii, and the line shows the LLM tokens required to reach ShinkaEvolve's final performance.}
\label{fig:ablation_sensitivity}

\vspace{8pt}
\centering
\includegraphics[width=\textwidth]{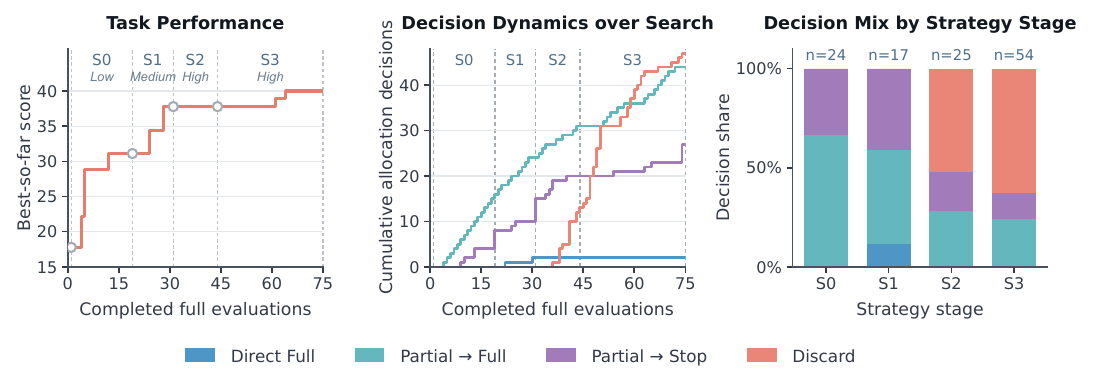}
\caption{\textbf{Evolution of allocation behavior on ADAS-AIME.} A representative 75-full-evaluation run showing best-so-far performance and allocation decisions across S0--S3. Allocation panels show decisions before counterfactual exploration, with explored proposals counted by their original allocation decision. Dashed lines mark the adoption of validated Strategies.}
\label{fig:adas_policy_evolution}
\end{figure}

\subsection{Main Results}
\label{sec:main_results}

Figure~\ref{fig:main_sample_efficiency} shows that EvoAlloc achieves the highest mean final performance among all
evaluated methods on both benchmarks.
Relative to ShinkaEvolve, the underlying program-evolution framework, EvoAlloc improves
mean final performance by $12.0$\% on ADAS-AIME and $8.7$\% on Circle Packing.
It also substantially reduces matched-target cost, requiring $59$--$82$\% fewer Full
evaluations and $61$--$89$\% fewer LLM tokens to reach ShinkaEvolve's final-performance
target.
LLM-token cost includes all input and output tokens consumed during search, including
proposal generation, benchmark evaluation, and method-specific allocation or selection calls.
The improvements persist across both evaluation settings: on Circle Packing, where no
Partial evaluation is available, EvoAlloc still benefits from selectively allocating Full
evaluations, showing that the gains are not simply due to access to an additional
evaluation stage.
The consistent advantage over the One-Step LLM Allocator further supports the value of
maintaining an explicit Strategy and accumulated Experience across the search rather than
making independent LLM allocation decisions.
Complete statistics, matched-target costs, and auxiliary trajectory AUC are reported in
Appendix~\ref{app:detailed_results}.

\subsection{Ablation and Sensitivity Analysis}
\label{sec:controlled_design_main}

Figure~\ref{fig:ablation_sensitivity}(a--b) compares variants of EvoAlloc.
\textbf{One-Step Alloc} replaces the maintained Strategy with independent one-step
allocation decisions, using the same allocator LLM and decision-time search context as EvoAlloc.
\textbf{w/o Evolution} uses fixed custom allocation guidance, distilled from baseline
run examples and manually refined (Appendix~\ref{app:baselines_ablations}).
\textbf{w/o Exploration} removes counterfactual exploration while leaving the remaining
adaptation mechanisms unchanged.
Across both benchmarks, the full EvoAlloc achieves the strongest mean best-so-far trajectories.
The full EvoAlloc outperforms both one-step allocation and fixed custom guidance.
Removing exploration also weakens performance, reflecting the value of counterfactual
outcomes that selective allocation would otherwise leave unobserved.
Figure~\ref{fig:ablation_sensitivity}(c) varies the exploration rate $\rho$ on Circle Packing. Final performance is similar across rates, but the LLM tokens needed to reach ShinkaEvolve's final performance vary substantially, with $\rho = 0.3$ the most efficient. Too little exploration provides limited feedback on skipped candidates, whereas too much offsets the savings from selective allocation.

\subsection{How the Allocation Strategy Evolves and Reshapes Search}
\label{sec:critic_evolution}
\label{sec:tree_analysis}

\paragraph{A good allocation strategy changes as search progresses.}
What is worth evaluating depends on the state of the search: early outcomes are hard to predict and gains are frequent, so broad evaluation is informative, whereas later gains become rare and the budget is better spent on fewer, stronger candidates. EvoAlloc's evolved strategies follow this shift (Figure~\ref{fig:adas_policy_evolution}; see Appendix~\ref{app:representative_guidance_circle} for Circle Packing). On ADAS-AIME, allocation moves from gathering evidence through Partial evaluation with no direct discards to discarding most proposals outright. This greater selectivity does not stop discovery, as EvoAlloc still finds a late improvement. These stage-wise changes support revising allocation criteria as search
experience accumulates.

\begin{figure}[!t]
\centering
\includegraphics[width=\textwidth]{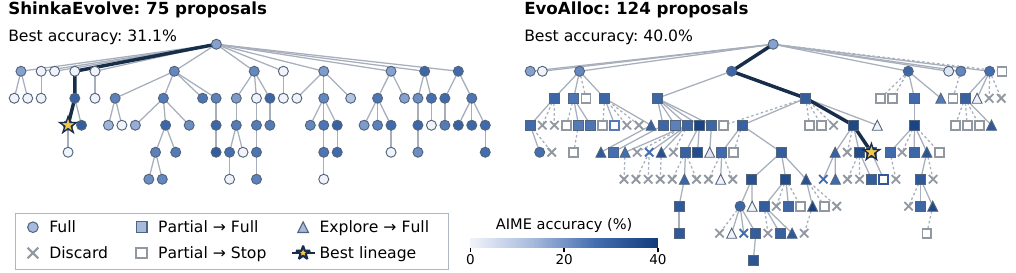}
\caption{\textbf{ADAS-AIME search trees.} Both methods complete $75$ full
evaluations with the same seed. Shapes encode allocation decisions; colors indicate
observed Full-evaluation accuracy. Gray markers denote proposals without Full
outcomes; stars mark the best programs, and highlighted edges trace their lineages.}
\label{fig:tree_visualization}
\vspace{-6pt}
\end{figure}

\begin{figure}[!t]
\centering
\includegraphics[width=\linewidth]{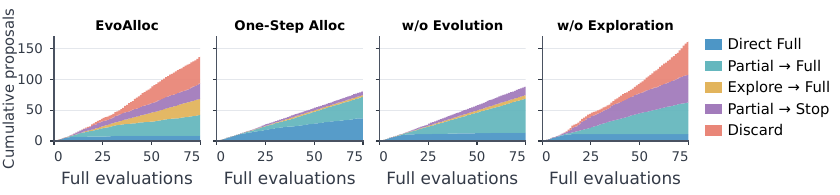}
\caption{\textbf{Allocation decisions under ablations on ADAS-AIME.} Five-seed mean cumulative proposal counts by allocation decision as Full evaluations accumulate; the stacked total gives the number of proposals considered.}
\label{fig:allocation_patterns}
\vspace{-10pt}
\end{figure}

\paragraph{Savings become exploration.}
Selective allocation does more than reduce cost: every evaluation withheld from a weak candidate is reinvested in further search. With the same Full-evaluation budget on the same seed, EvoAlloc examines 124 proposals on ADAS-AIME versus 75 for ShinkaEvolve (Figure~\ref{fig:tree_visualization}), and 302 versus 150 on Circle Packing (Appendix~\ref{app:tree_visualization}). The extra proposals expand the tree with additional siblings and descendants. The candidates EvoAlloc chooses to evaluate also tend to score higher, even though each choice is made before the Full outcome is known, so the broader coverage does not come at the expense of evaluation quality. This reinvestment is why EvoAlloc improves final performance under a fixed budget, not only the cost of reaching a given target.

\paragraph{EvoAlloc components jointly shape allocation behavior.}
Figure~\ref{fig:allocation_patterns} shows how removing each component changes allocation behavior on ADAS-AIME. One-Step Alloc sees the same search history, but without a Strategy that consolidates it into explicit criteria, the LLM cannot confidently decline, and it therefore sends most proposals to Full evaluation and saves little budget. With fixed custom guidance, allocation remains Partial-heavy throughout the search, unlike the stage-wise shifts of EvoAlloc. Without exploration, discarded outcomes stay hidden, leaving no counterexamples to the current criteria, so pruning errors go uncorrected and falsely pruned candidates never enter the archive as parents. Effective allocation thus requires explicit criteria, their evolution with the search, and exploration to check them against hidden outcomes.

\section{Conclusion and Limitations}
\label{sec:conclusion}

We introduced EvoAlloc, a self-evolving agent that treats evaluation allocation as an evolving component of program evolution. EvoAlloc revises an explicit allocation Strategy from accumulated Experiences and uses counterfactual exploration to recover feedback that selective allocation would otherwise leave hidden. Across ADAS-AIME and Circle Packing, EvoAlloc improves final performance and reaches the reference final-performance targets with $59$--$82$\% fewer Full evaluations and $61$--$89$\% fewer LLM tokens. These findings point toward self-improving systems that learn not only what to search, but also where to spend their evaluation effort.

Our study examines self-evolving allocation at the evaluation layer on two benchmarks with contrasting evaluators, within a single search run and with generation, evaluators, and update mechanisms held fixed. Extending resource allocation to generation and allowing evaluation and update mechanisms to evolve could broaden the scope of self-improvement \citep{sia,dgm}. Retaining allocation experience across runs could allow future allocators to build on previously learned strategies, supporting longer-term knowledge accumulation and inheritance, as learngenes do by passing condensed knowledge across generations of agents \citep{feng2025learngene}. More broadly, adaptive resource allocation could help systems pursuing open-ended self-improvement continually refine how they invest finite computation in future progress.

\clearpage

\section*{AI Use Statement}
Generative AI tools were used to assist with code implementation and debugging,
as well as language editing and presentation of the manuscript. The core research
idea and experimental design were developed by the authors. AI-assisted code was
reviewed and tested by the authors before use in the experiments, and all
AI-assisted manuscript content was manually reviewed and revised for technical
correctness and consistency with the reported results. The authors take full
responsibility for the final content of the paper.

\section*{Reproducibility Statement}
We provide detailed information to support reproducibility in the appendix,
including the EvoAlloc pseudocode, implementation details and hyperparameters,
benchmark and evaluation protocols, budget accounting, and all prompts required
to reproduce the method. We additionally report five-seed results and allocation
breakdowns to facilitate verification of the main findings. The code and
configurations will be released upon acceptance.

\bibliographystyle{iclr2027_conference}
\bibliography{refs}

\clearpage
\appendix
\raggedbottom
\section*{Appendix}
\input{sections/appendix}
\clearpage
\input{sections/artifacts_prompts}
\end{document}

%% file: math_commands.tex
\usepackage{amsmath,amsfonts,bm}

\def\eqref#1{equation~\ref{#1}}

\def\1{\bm{1}}

\newcommand{\test}{\mathcal{D_{\mathrm{test}}}}

\DeclareMathAlphabet{\mathsfit}{\encodingdefault}{\sfdefault}{m}{sl}
\SetMathAlphabet{\mathsfit}{bold}{\encodingdefault}{\sfdefault}{bx}{n}



%% file: sections/related_work.tex
\section{Related Work}
\label{sec:related_work}

\noindent\textbf{Program Evolution and Self-Improving Search.}
Program evolution has a long history in genetic programming, where evolutionary
variation and selection are applied directly to executable
programs~\citep{cramer1985representation,dickmanns1987genetische}.
LLM-based systems extend this paradigm by using language models to generate,
revise, and recombine candidate programs within iterative
search~\citep{funsearch,novikov2025alphaevolve}.
Recent work has further made the search process itself more adaptive through
mechanisms such as parent selection, novelty control, population management,
and learned search strategies~\citep{lange2025shinkaevolve,assumpcao2025codeevolve,wan2025loongflow}.
A related line of work considers self-improvement beyond the candidate programs
themselves. Early meta-evolution explored evolving program-modification
mechanisms~\citep{schmidhuber1987}, while the G\"odel Machine formalized
self-improvement as rewriting its own code once a modification is proven
useful~\citep{godelmachine}. More recent systems such as Darwin G\"odel Machine
and Huxley-G\"odel Machine pursue this idea empirically by evolving coding
agents and their problem-solving procedures~\citep{dgm,hgm}.
While these approaches largely adapt how candidates are generated and searched, the evaluation
mechanism that allocates compute typically remains fixed; EvoAlloc allows this mechanism to
evolve as well, drawing on accumulated search experience.

\noindent\textbf{Selective Evaluation and Adaptive Resource Allocation.}
Efficient search depends on how computation is allocated, particularly when candidate evaluation is expensive.
Earlier work on online algorithm portfolios adaptively allocated computation among a given set of search algorithms or parameter configurations, using performance prediction or bandit-based methods to adjust resource shares online~\citep{Gagliolo:04ecml, Gagliolo:05icann, DBLP:conf/dcai/GaglioloS08}.
A complementary line of work allocates evaluation effort across candidate solutions: surrogate-assisted methods predict candidate quality to reduce expensive evaluations~\citep{snoek2012bo,jin2011saea}, while multi-fidelity methods progressively allocate resources using partial evaluations~\citep{li2018hyperband,li2020asha}.
Racing and adaptive capping discard unpromising configurations or terminate runs early~\citep{hutter2009paramils,lopez2016irace}.
In LLM-based program evolution, selective evaluation takes the form of cascade evaluation through increasingly comprehensive tests~\citep{novikov2025alphaevolve}, novelty-based rejection of near-duplicate proposals~\citep{lange2025shinkaevolve}, and LLM-based candidate ranking or filtering before costly evaluation~\citep{foster2026rpm}.
These program-evolution approaches can adapt individual decisions to available evidence while retaining predefined allocation rules.
EvoAlloc instead revises an explicit allocation strategy using accumulated search and evaluation outcomes.

%% file: sections/appendix.tex
\captionsetup[table]{font=normalsize,justification=justified,singlelinecheck=false,format=plain,indention=0pt,skip=4pt}
\newcommand{\appendixtablecaption}[3]{%
  \refstepcounter{table}\label{#1}%
  \noindent{Table~\thetable: \textbf{#2} #3}\par\vspace{4pt}%
}

\section{EvoAlloc Algorithm}
\label{app:pseudocode}

\input{sections/algorithm}

\clearpage
\section{Implementation and Benchmarks}
\label{app:impl}
\label{app:benchmark_programs}

\subsection{Models and EvoAlloc configuration}
\label{app:models_configuration}

EvoAlloc is implemented on top of the public ShinkaEvolve program-evolution
loop~\citep{lange2025shinkaevolve}. Unless otherwise stated, substrate
hyperparameters follow the default ShinkaEvolve configuration.
All language models are open-weight and locally served on NVIDIA A100
GPUs, with the same serving setup used across all compared methods.
Table~\ref{tab:appendix_impl_summary} summarizes the backbone models, the main
EvoAlloc configuration, and the benchmark-specific settings.

\vspace{0.7em}
\begingroup
\appendixtablecaption{tab:appendix_impl_summary}{Implementation and benchmark configuration.}{Shared model and controller settings with benchmark-specific budgets, action spaces, and Partial-evaluation availability.}
\small
\setlength{\tabcolsep}{5pt}
\renewcommand{\arraystretch}{1.00}
\noindent\begin{tabular}{@{}>{\raggedright\arraybackslash}p{0.35\textwidth}@{}>{\raggedright\arraybackslash}p{0.25\textwidth}@{}>{\raggedright\arraybackslash}p{0.4\textwidth}@{}}
\toprule
\multicolumn{3}{@{}l}{\textbf{Backbone and Models}} \\
\midrule
Proposer LLMs & \multicolumn{2}{@{}>{\raggedright\arraybackslash}p{0.65\textwidth}@{}}{GLM-4.5-Air; gpt-oss-120b; Qwen3.6-35B-A3B} \\
ShinkaEvolve meta model & \multicolumn{2}{@{}>{\raggedright\arraybackslash}p{0.65\textwidth}@{}}{Qwen3.6-35B-A3B} \\
ShinkaEvolve novelty model & \multicolumn{2}{@{}>{\raggedright\arraybackslash}p{0.65\textwidth}@{}}{Qwen3.6-35B-A3B} \\
Embedding model & \multicolumn{2}{@{}>{\raggedright\arraybackslash}p{0.65\textwidth}@{}}{Qwen3-Embedding-0.6B} \\
Proposer temperatures & \multicolumn{2}{@{}>{\raggedright\arraybackslash}p{0.65\textwidth}@{}}{$[0.0,\,0.5,\,1.0]$} \\
Max tokens & \multicolumn{2}{@{}>{\raggedright\arraybackslash}p{0.65\textwidth}@{}}{16,384} \\
\midrule
\multicolumn{3}{@{}l}{\textbf{EvoAlloc Configuration}} \\
\midrule
Allocator and self-evolution LLM & \multicolumn{2}{@{}>{\raggedright\arraybackslash}p{0.65\textwidth}@{}}{Qwen3-Coder-Next} \\
Adaptive warm-up & \multicolumn{2}{@{}>{\raggedright\arraybackslash}p{0.65\textwidth}@{}}{Requires \(\geq 1\) new-best and \(\geq 1\) non-new-best outcome} \\
Experience update interval $k_E$ & \multicolumn{2}{@{}>{\raggedright\arraybackslash}p{0.65\textwidth}@{}}{5 new Full outcomes} \\
Strategy update interval $k_S$ & \multicolumn{2}{@{}>{\raggedright\arraybackslash}p{0.65\textwidth}@{}}{10 new Full outcomes} \\
Strategy validation window & \multicolumn{2}{@{}>{\raggedright\arraybackslash}p{0.65\textwidth}@{}}{2 allocation disagreements with known Full outcomes} \\
Experience capacity & \multicolumn{2}{@{}>{\raggedright\arraybackslash}p{0.65\textwidth}@{}}{4 active/tentative + 4 stale} \\
Counterfactual exploration rate & \multicolumn{2}{@{}>{\raggedright\arraybackslash}p{0.65\textwidth}@{}}{$0.30$} \\
\end{tabular}\par\nointerlineskip
\noindent\begin{tabular}{@{}>{\raggedright\arraybackslash}p{0.25\textwidth}@{}>{\raggedright\arraybackslash}p{0.30\textwidth}@{}>{\raggedright\arraybackslash}p{0.45\textwidth}@{}}
\midrule
\multicolumn{3}{@{}l}{\textbf{Benchmark-Specific Configuration}} \\
\midrule
& \textbf{Circle Packing} & \textbf{ADAS-AIME} \\
Objective / metric & Packing score & AIME accuracy \\
Evaluation model & Deterministic evaluator & Qwen3-8B solver (thinking off) \\
Full-evaluation budget & 150 & 75 \\
Allocator actions & $\{\texttt{FULL\_EVAL},\,\texttt{DISCARD}\}$ & $\{\texttt{FULL\_EVAL},\,\texttt{PARTIAL\_EVAL},\,\texttt{DISCARD}\}$ \\
Partial-evaluation subset & --- & Fixed subset, \(k=8\) (first 8 of 30 problems) \\
Post-Partial actions & --- & $\{\texttt{CONTINUE\_TO\_FULL},\,\texttt{STOP}\}$ \\
\bottomrule
\end{tabular}
\endgroup

\paragraph{Strategy initialization.}
We initialize $S_0$ by summarizing 50 example cases, comprising
10 candidate evaluation records from each of five ShinkaEvolve baseline
runs with different seeds. The initial selectivity is set to \texttt{low}.
The cost of obtaining these baseline cases is excluded from the reported
search costs.

\clearpage
\input{sections/benchmark_protocols}

\clearpage
\section{Evolution and Search Analysis}
\label{app:representative_guidance}

This appendix examines how EvoAlloc's Strategy and Experiences evolve during search, and how these changes reshape the search itself. We first trace their evolution on ADAS-AIME and Circle Packing, then analyze the resulting search trees.

\subsection{ADAS-AIME Strategy and Experience Evolution}
\label{app:representative_guidance_adas}

\paragraph{Strategy trajectory.}
Figure~\ref{fig:adas_policy_evolution} visualizes the representative ADAS-AIME run analyzed here. The adopted Strategy starts permissive, moves to medium selectivity once sufficient evidence has accumulated, and then passes through two highly selective stages. Its criteria also become more specific over time: the generic guidance of S0 gives way to explicit Partial-evaluation thresholds conditioned on lineage trends. Table~\ref{tab:adas_representative_policy} lists each stage's Full-evaluation window, selectivity level, and complete Strategy text; the final Experience inventory is reported afterward.

\vspace{0.7em}
\begingroup
\appendixtablecaption{tab:adas_representative_policy}{Evolution of the learned allocation Strategy on ADAS-AIME.}{Each row lists one Strategy stage with its full-evaluation window, selectivity level, and complete Strategy text, with ``smoke'' replaced by ``partial evaluation'' for consistency with the main text.}
\small
\setlength{\tabcolsep}{4pt}
\renewcommand{\arraystretch}{1.16}
\noindent\begin{tabularx}{\textwidth}{@{}ccc>{\raggedright\arraybackslash}X@{}}
\toprule
\textbf{Stage} & \textbf{Full window} & \textbf{Selectivity} & \textbf{Allocation strategy} \\
\midrule
S0
& 1--19
& Low
& Allocate evaluator calls using the available evidence while preserving useful
uncertainty. Preserve evaluator calls without hiding promising or informative
candidates. \\
\addlinespace

S1
& 20--31
& Medium
& Prefer direct full evaluation for structural or prompting redesigns with at least 3 of 8 partial-evaluation successes. Use partial evaluation for parameter-level changes or unclear lineages, and reserve discarding for poor partial-evaluation execution or lineages already at the current best without a novel mechanism. \\
\addlinespace

S2
& 32--44
& High
& Reserve full evaluation for strong structural redesigns with at least 4 of 8 partial-evaluation successes, especially when the lineage is improving or the parent is near the current best. Continue after partial evaluation only under the same evidence; otherwise stop early, and discard proposals with few partial-evaluation successes from flat or regressive lineages more aggressively. \\
\addlinespace

S3
& 45--75
& High
& Reserve full evaluation for structural redesigns with at least 4 of 8 partial-evaluation successes and clear lineage improvement. Use partial evaluation for parameter-level or unclear-lineage proposals, but continue only with both strong partial-evaluation evidence and an improving lineage. Aggressively discard proposals with at most 2 of 8 partial-evaluation successes when the lineage is flat or declining or the change lacks structural novelty. \\
\bottomrule
\end{tabularx}
\endgroup
\vspace{0.7em}

\clearpage

\paragraph{Final learned Experiences.}
The Experience inventory records recurring search situations and the outcome tendencies distilled from them during the run. Each entry specifies an applicability condition, an expected tendency, and its supporting and contradicting evidence. Unlike the run-level Strategy, Experiences retain local, case-grounded information that can be reused when similar search states reappear. Conditions that refer to Partial-evaluation results apply only after Partial evaluation has been observed. As in Table~\ref{tab:adas_representative_policy}, ``smoke'' in the original text is replaced by ``partial evaluation''.

\begin{experiencebox}{Experience 1\hfill\textit{\color{triton_green!80!black}active}}
\experiencefield{triton_green!80!black}{APPLICABILITY}{Proposals with at most 2
of 8 partial-evaluation successes, especially when the lineage is flat or declining, even
when the proposed structural change appears promising.}
\experiencefield{triton_green!80!black}{EXPECTED TENDENCY}{They tend to yield
low or non-improving full scores regardless of design ambition.}
\experiencefield{triton_green!85!black}{EVIDENCE}{15 support / 3 counter.}
\end{experiencebox}

\begin{experiencebox}{Experience 2\hfill\textit{\color{triton_green!80!black}active}}
\experiencefield{triton_green!80!black}{APPLICABILITY}{Targeted deterministic
solvers with at least 4 of 8 partial-evaluation successes when the lineage is improving and
the parent is at or near the current best.}
\experiencefield{triton_green!80!black}{EXPECTED TENDENCY}{They can improve over
the parent and sometimes establish a new best when the fix targets a known
failure mode.}
\experiencefield{triton_green!85!black}{EVIDENCE}{2 support / 0 counter.}
\end{experiencebox}

\begin{experiencebox}[colback=black!2,colframe=black!18,colbacktitle=black!12]{Experience 3\hfill\textit{\color{black!55}stale}}
\experiencefield{black!55}{APPLICABILITY}{Targeted deterministic solvers for
recurring AIME patterns with at most 2 of 8 partial-evaluation successes, especially on a
flat or declining lineage.}
\experiencefield{black!55}{EXPECTED TENDENCY}{They may still improve over the
parent when the structural fix targets a known blind spot.}
\experiencefield{black!55}{EVIDENCE}{2 support / 2 counter.}
\end{experiencebox}

\begin{experiencebox}[colback=black!2,colframe=black!18,colbacktitle=black!12]{Experience 4\hfill\textit{\color{black!55}stale}}
\experiencefield{black!55}{APPLICABILITY}{Targeted deterministic solvers with
at least 4 of 8 partial-evaluation successes, especially on a flat or declining lineage.}
\experiencefield{black!55}{EXPECTED TENDENCY}{They often plateau at or below
the current best despite strong partial-evaluation results and structural novelty.}
\experiencefield{black!55}{EVIDENCE}{4 support / 1 counter.}
\end{experiencebox}

\clearpage

\subsection{Circle Packing Strategy and Experience Evolution}
\label{app:representative_guidance_circle}

\begin{figure}[H]
\centering
\includegraphics[width=\textwidth]{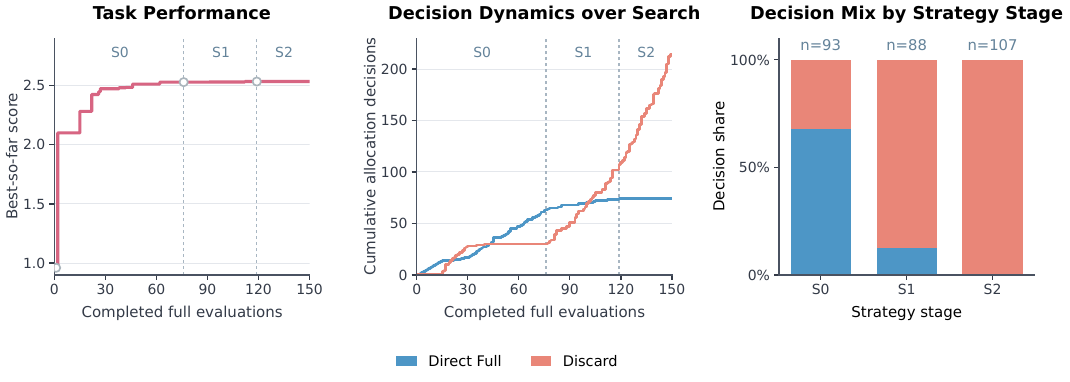}
\caption{\textbf{Evolution of allocation behavior on Circle Packing.} A representative 150-full-evaluation run (seed 3) showing best-so-far performance and allocation decisions across S0--S2. Allocation panels show decisions before counterfactual exploration, with explored proposals counted by their original allocation decision. Dashed lines mark the adoption of validated Strategies.}
\label{fig:circle_policy_evolution_appendix}
\end{figure}

\paragraph{Strategy trajectory.}
Figure~\ref{fig:circle_policy_evolution_appendix} shows a representative Circle Packing run. Its Strategy follows a progression in selectivity similar to ADAS-AIME, but without Partial evaluation, so each decision is made directly between Full evaluation and discard. The allocator first evaluates proposals broadly, then becomes increasingly selective as evidence accumulates and the search matures. Table~\ref{tab:circle_seed3_policy} lists the corresponding Strategy stages.

\begingroup
\appendixtablecaption{tab:circle_seed3_policy}{Evolution of the learned allocation Strategy on Circle Packing.}{Each row lists one Strategy stage with its full-evaluation window, selectivity level, and complete Strategy text.}
\small
\setlength{\tabcolsep}{4pt}
\renewcommand{\arraystretch}{1.16}
\noindent\begin{tabularx}{\textwidth}{@{}ccc>{\raggedright\arraybackslash}X@{}}
\toprule
\textbf{Stage} & \textbf{Full window} & \textbf{Selectivity} & \textbf{Allocation strategy} \\
\midrule
S0
& 1--76
& Low
& Begin with low selectivity. Early proposal outcomes are highly variable, and
surface plausibility does not reliably distinguish failures from improvements.
Evaluate broadly until grounded outcomes establish a repeatable basis for
discarding, then tighten allocation gradually. Use the limited evaluation budget
efficiently over the full search. \\
\addlinespace

S1
& 77--119
& Medium
& Begin to favor targeted evaluation. Retain broad coverage for proposals with
strong lineage improvement trends or structural novelty, but increasingly
discard repeated marginal variants or invalid proposals without structural
divergence. Prioritize full evaluation for candidates that extend a promising
new-best lineage or introduce a substantially new initialization or
optimization strategy. \\
\addlinespace

S2
& 120--150
& High
& Strongly favor discarding repeated marginal variants on plateaued lineages
or motifs associated with invalid and non-improving cases. Reserve full
evaluation primarily for structurally novel strategies with promising lineage
evidence, or for candidates directly extending a new-best lineage and plausibly
escaping a local optimum. \\
\bottomrule
\end{tabularx}
\endgroup

\clearpage

\paragraph{Final learned Experiences.}
The final inventory summarizes recurring proposal patterns and their observed
outcomes. Descriptions such as ``valid but non-improving'' characterize past
evaluated cases, not conditions to be checked on a new candidate.
These Experiences inform Strategy revision and expectations about related
proposals through visible code changes and known lineage outcomes.
Allocation uses only information available at decision time, without access
to the current candidate's Full outcome (Appendix~\ref{app:allocator_prompt}).

\begin{experiencebox}{Experience 1\hfill\textit{\color{triton_green!80!black}active}}
\experiencefield{triton_green!80!black}{APPLICABILITY}{Valid but substantially
non-improving candidates, especially minor perturbations of invalid or
low-quality structures.}
\experiencefield{triton_green!80!black}{EXPECTED TENDENCY}{Such candidates rarely
improve without structural innovation.}
\experiencefield{triton_green!85!black}{EVIDENCE}{9 support / 0 counter.}
\end{experiencebox}

\begin{experiencebox}{Experience 2\hfill\textit{\color{triton_green!80!black}active}}
\experiencefield{triton_green!80!black}{APPLICABILITY}{Invalid outputs associated with
structural flaws, such as overlap or boundary violations, often following
aggressive or poorly constrained edits.}
\experiencefield{triton_green!80!black}{EXPECTED TENDENCY}{Invalid outcomes rarely
precede improvement without corrective refinement.}
\experiencefield{triton_green!85!black}{EVIDENCE}{28 support / 0 counter.}
\end{experiencebox}

\begin{experiencebox}{Experience 3\hfill\textit{\color{triton_green!80!black}active}}
\experiencefield{triton_green!80!black}{APPLICABILITY}{Candidates that improve on a
valid parent, even when their absolute quality remains low.}
\experiencefield{triton_green!80!black}{EXPECTED TENDENCY}{Local parent improvement
can reveal useful structural motifs and may precede larger gains after further
refinement.}
\experiencefield{triton_green!85!black}{EVIDENCE}{58 support / 9 counter.}
\end{experiencebox}

\begin{experiencebox}[colback=black!2,colframe=black!18,colbacktitle=black!12]{Experience 4\hfill\textit{\color{black!55}stale}}
\experiencefield{black!55}{APPLICABILITY}{Rare new-best outcomes that emerge
after repeated invalid or low-scoring attempts.}
\experiencefield{black!55}{EXPECTED TENDENCY}{Such outcomes are not reliably
predictable from prior local scores and may require full evaluation to confirm.}
\experiencefield{black!55}{EVIDENCE}{9 support / 1 counter.}
\end{experiencebox}
\clearpage
\subsection{Search-Tree Analysis}
\label{app:additional_search_trajectory}
\label{app:tree_visualization}

\begin{figure}[H]
\centering
\includegraphics[width=\textwidth]{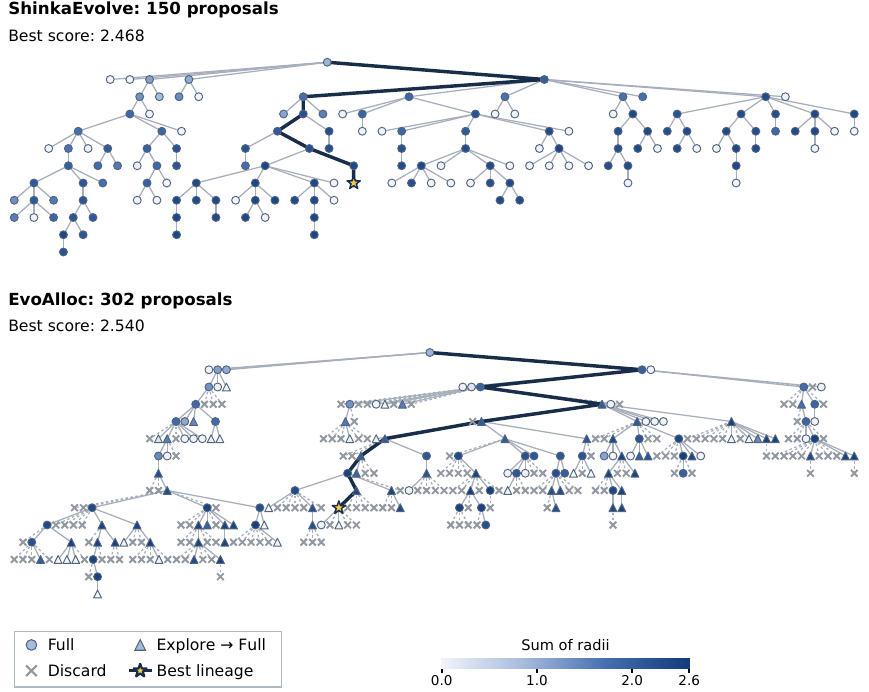}
\caption{\textbf{Circle Packing search trees.} Both methods complete $150$ full
evaluations with the same seed (seed 1). Shapes encode allocation decisions; colors indicate
the observed sum of radii. Gray crosses denote proposals without Full outcomes;
stars mark the best programs, and highlighted edges trace their lineages.}
\label{fig:tree_circle_appendix}
\end{figure}

Figure~\ref{fig:tree_circle_appendix} complements Section~\ref{sec:tree_analysis} with Circle Packing search trees. With the same seed and 150 Full evaluations, EvoAlloc considers 302 proposals, compared with 150 for ShinkaEvolve. Since Circle Packing has no Partial evaluation, this expansion comes entirely from deciding which proposals receive Full evaluation. Highlighted edges trace the lineages of the best programs.

\clearpage
\input{sections/presubmission_checklist}

\subsection{Detailed Performance}
\label{app:detailed_performance}

Table~\ref{tab:quantitative-performance} reports final performance and normalized trajectory AUC as an auxiliary anytime measure; matched-target cost remains the primary efficiency metric. For each seed, we normalize the area under the best-so-far curve by the range length, using the full evaluation budget for eval-budget AUC and the common token range covered by all methods and seeds for token-budget AUC ($0.62$--$55.1$ M tokens on ADAS-AIME and $0.27$--$4.68$ M on Circle Packing). We report the mean and standard deviation across five seeds, with standard deviations computed under the population convention to match the shaded bands in Figure~\ref{fig:main_sample_efficiency}.

\definecolor{targetshade}{RGB}{244,245,247}
\newcommand{\ours}[1]{\textbf{\boldmath #1}}
\newcommand{\meanstd}[2]{$#1\,{\scriptstyle\pm #2}$}
\newcommand{\boldmeanstd}[2]{$\mathbf{#1}\,{\scriptstyle\mathbf{\pm #2}}$}

\begin{table}[H]
\centering
\footnotesize
\caption{\textbf{Final performance and auxiliary trajectory AUC.} Five-seed mean $\pm$ one standard deviation; bold denotes the best method per benchmark and column. AUC is the normalized area under the best-so-far curve over the evaluation or token budget; higher is better.}
\label{tab:quantitative-performance}
\setlength{\tabcolsep}{5pt}
\renewcommand{\arraystretch}{1.18}
\begin{tabular*}{\textwidth}{@{\extracolsep{\fill}}ll ccc@{}}
\toprule
\textbf{Benchmark} & \textbf{Method} & \textbf{Performance}
& \textbf{Eval-budget AUC} & \textbf{Token-budget AUC} \\
\midrule
\multirow{5}{*}{ADAS-AIME} & ShinkaEvolve & \meanstd{35.33}{2.57} & \meanstd{31.29}{1.95} & \meanstd{30.24}{2.82} \\
 & One-Step Alloc & \meanstd{34.89}{1.94} & \meanstd{31.40}{1.34} & \meanstd{32.45}{1.49} \\
 & Lineage-UCB & \meanstd{34.00}{0.89} & \meanstd{30.88}{0.53} & \meanstd{32.32}{1.07} \\
 & RPM & \meanstd{36.22}{1.13} & \meanstd{32.49}{0.69} & \meanstd{30.31}{0.94} \\
 & \ours{EvoAlloc (ours)} & \boldmeanstd{39.56}{6.80} & \boldmeanstd{33.69}{3.33} & \boldmeanstd{33.15}{2.89} \\
\midrule
\multirow{5}{*}{Circle Packing} & ShinkaEvolve & \meanstd{2.343}{0.254} & \meanstd{2.172}{0.239} & \meanstd{2.245}{0.257} \\
 & One-Step Alloc & \meanstd{2.429}{0.219} & \meanstd{2.261}{0.222} & \meanstd{2.251}{0.266} \\
 & Lineage-UCB & \meanstd{2.408}{0.131} & \meanstd{2.181}{0.156} & \meanstd{2.213}{0.147} \\
 & RPM & \meanstd{2.390}{0.186} & \meanstd{2.246}{0.151} & \meanstd{1.553}{0.277} \\
 & \ours{EvoAlloc (ours)} & \boldmeanstd{2.547}{0.030} & \boldmeanstd{2.389}{0.043} & \boldmeanstd{2.436}{0.024} \\
\bottomrule
\end{tabular*}
\end{table}

\subsection{Resource and Token Costs}
\label{app:resource_costs}

Table~\ref{tab:sample-efficiency} summarizes the Full evaluations and total LLM tokens needed to reach matched performance targets. For each method, we align the five seed trajectories by completed Full evaluations and average both their best-so-far scores and cumulative token usage at each evaluation index. We then identify the first index at which the mean score reaches the target. The reported costs are this Full-evaluation index and the corresponding mean cumulative token usage.

Target matching is therefore performed after averaging across seeds, rather than averaging each seed's individual first-hit cost. In particular, the reported token cost is not the first-hit token budget of the token-aligned mean performance curve shown in Figure~\ref{fig:main_sample_efficiency}. It is the mean cumulative token usage at the target-reaching Full-evaluation index, as used consistently in Table~\ref{tab:sample-efficiency} and the bottom panels of Figure~\ref{fig:main_sample_efficiency}.

The reference final target is ShinkaEvolve's five-seed mean final score. Let $Q_{\mathrm{initial}}$ and $Q_{\mathrm{final}}$ denote the first and final values of its mean best-so-far curve. The 95\%-gain target is
\begin{equation}
  q_{95} = Q_{\mathrm{initial}} + 0.95\,(Q_{\mathrm{final}} - Q_{\mathrm{initial}}),
  \label{eq:app-95-gain-target}
\end{equation}
corresponding to 95\% of ShinkaEvolve's improvement over its initial score.

\paragraph{Budget accounting.}
All Full evaluations performed during online search count toward the
same search budget, including those from warm-up, counterfactual
exploration, and Strategy validation; shared evaluations are counted once. Recorded LLM usage includes proposal generation, benchmark evaluation, and allocator calls for resource allocation, self-evolution, and shadow validation, and Partial evaluations are counted whether a proposal stops or continues. For RPM, allocator costs are the tokens spent on pairwise comparisons; Lineage-UCB requires no LLM calls for allocation. Table~\ref{tab:resource-accounting} reports costs at the full evaluation budget. At this budget, EvoAlloc consumes more total tokens than ShinkaEvolve, since the evaluations it withholds are reinvested in examining additional proposals (Section~\ref{sec:tree_analysis}). Its allocator accounts for $1.02$ M tokens on ADAS-AIME and $1.67$ M on Circle Packing. Total usage need not scale linearly with the number of proposals, as generated programs differ in their evaluation costs. The savings in Table~\ref{tab:sample-efficiency} therefore reflect how quickly each method reaches a given performance level, rather than its total consumption once the budget is exhausted.

\begin{table}[H]
\centering
\footnotesize
\caption{\textbf{Resources to matched performance targets.} First-hit costs
read from the five-seed mean best-so-far curves, with mean cumulative token
usage taken at the corresponding Full-evaluation index. Parentheses show
changes relative to ShinkaEvolve; dashes indicate that the mean curve does not
reach the target within the observed evaluation budget.}
\label{tab:sample-efficiency}
\setlength{\tabcolsep}{4.5pt}
\renewcommand{\arraystretch}{1.16}
\begin{tabularx}{\textwidth}{@{}l *{4}{>{\centering\arraybackslash}X}@{}}
\toprule
& \multicolumn{2}{c}{\textbf{Reference final target (100\%)}}
& \multicolumn{2}{c}{\textbf{95\%-gain target}} \\
\cmidrule(lr){2-3}\cmidrule(l){4-5}
\textbf{Method}
& \mbox{\textbf{Full evals} $\downarrow$}
& \mbox{\textbf{LLM tokens (M)} $\downarrow$}
& \mbox{\textbf{Full evals} $\downarrow$}
& \mbox{\textbf{LLM tokens (M)} $\downarrow$} \\
\midrule
\rowcolor{black!5}
\textbf{ADAS-AIME} & \multicolumn{2}{c}{target: $\geq 35.33$} & \multicolumn{2}{c}{target: $\geq 34.49$} \\
ShinkaEvolve & $69$ & $89.8$ & $51$ & $63.6$ \\
One-Step Alloc & -- & -- & $66$ ($+29\%$) & $77.7$ ($+22\%$) \\
Lineage-UCB & -- & -- & -- & -- \\
RPM & $66$ ($-4\%$) & $140.1$ ($+56\%$) & $57$ ($+12\%$) & $117.2$ ($+84\%$) \\
\ours{EvoAlloc (ours)} & \ours{$28$ ($-59\%$)} & \ours{$35.3$ ($-61\%$)} & \ours{$26$ ($-49\%$)} & \ours{$32.7$ ($-49\%$)} \\
\midrule
\rowcolor{black!5}
\textbf{Circle Packing} & \multicolumn{2}{c}{target: $\geq 2.343$} & \multicolumn{2}{c}{target: $\geq 2.274$} \\
ShinkaEvolve & $148$ & $6.1$ & $70$ & $2.0$ \\
One-Step Alloc & $84$ ($-43\%$) & $5.0$ ($-18\%$) & $44$ ($-37\%$) & $1.9$ ($-4\%$) \\
Lineage-UCB & $95$ ($-36\%$) & $3.5$ ($-42\%$) & $76$ ($+9\%$) & $2.7$ ($+36\%$) \\
RPM & $67$ ($-55\%$) & $41.4$ ($+582\%$) & $40$ ($-43\%$) & $19.0$ ($+860\%$) \\
\ours{EvoAlloc (ours)} & \ours{$26$ ($-82\%$)} & \ours{$0.7$ ($-89\%$)} & \ours{$24$ ($-66\%$)} & \ours{$0.6$ ($-70\%$)} \\
\bottomrule
\end{tabularx}
\end{table}

\subsection{Allocation Breakdown}
\label{app:allocation_breakdown}

Table~\ref{tab:resource-accounting} shows how each method spends the same Full-evaluation budget. At this budget, EvoAlloc considers $136.8$ proposals per seed on ADAS-AIME and $243.0$ on Circle Packing, compared with $75.0$ and $150.0$ for ShinkaEvolve. Proposal counts include all candidates that reach the allocation stage and complete their allocation episode, with initialization and warm-up candidates counted as Direct Full. Allocation percentages are computed per seed over the same candidates and record each candidate's original decision rather than its evaluation outcome: candidates evaluated for Strategy validation remain in their original category, and Explore$\to$Full counts only candidates revealed by counterfactual exploration. For RPM, we report the nominal counts prescribed by its protocol, with 15 proposals per Full evaluation of which one is selected, while its token costs are measured from the actual runs.

\begin{table}[!htbp]
\centering
\footnotesize
\caption{\textbf{Token costs and allocation decisions.} Five-seed averages at the full evaluation budget; RPM allocation statistics follow its nominal batch protocol (Section~\ref{app:allocation_breakdown}). Allocator tokens include resource allocation, self-evolution, and shadow validation; for RPM, they are the tokens spent on pairwise selection. Dashes denote methods without an LLM-based allocator or allocation paths a method does not use. Allocation categories are mutually exclusive and sum to $100$\% within each method, up to rounding, with percentages computed per seed and then averaged. Percentages describe allocator decisions rather than all executed evaluations; candidates receiving additional Full evaluations during
validation are counted under their original decision categories.}
\label{tab:resource-accounting}
\setlength{\tabcolsep}{2pt}
\renewcommand{\arraystretch}{1.20}
\begin{tabular*}{\textwidth}{@{\extracolsep{\fill}}l ccc ccccc@{}}
\toprule
& \multicolumn{3}{c}{\textbf{Token \& proposal cost}}
& \multicolumn{5}{c}{\textbf{Allocation breakdown (\%)}} \\
\cmidrule(lr){2-4}\cmidrule(l){5-9}
\textbf{Method}
& \shortstack{\textbf{Total}\\\textbf{tokens (M)}}
& \shortstack{\textbf{Allocator/selector}\\\textbf{(M; \% of total)}}
& \shortstack{\textbf{Proposal}\\\textbf{count}}
& \shortstack{\textbf{Direct}\\\textbf{Full}}
& \shortstack{\textbf{Partial}\\$\rightarrow$\textbf{Full}}
& \shortstack{\textbf{Partial}\\$\rightarrow$\textbf{Stop}}
& \shortstack{\textbf{Direct}\\\textbf{Discard}}
& \shortstack{\textbf{Explore}\\$\rightarrow$\textbf{Full}} \\
\midrule
\rowcolor{black!5}\multicolumn{9}{@{}l}{\textbf{ADAS-AIME}} \\
ShinkaEvolve & $98.44$ & -- & $75.0$ & $100.0$ & -- & -- & -- & -- \\
One-Step Alloc & $92.53$ & $0.03$ ($<0.1\%$) & $81.2$ & $45.7$ & $43.4$ & $7.5$ & $0.0$ & $3.4$ \\
Lineage-UCB & $84.93$ & -- & $95.4$ & $78.7$ & -- & -- & $21.3$ & -- \\
RPM & $178.27$ & $10.98$ ($6.2$\%) & $1125.0$ & $6.7$ & -- & -- & $93.3$ & -- \\
\ours{EvoAlloc (ours)} & $128.76$ & $1.02$ ($0.8$\%) & $136.8$ & $6.2$ & $26.7$ & $19.1$ & $29.2$ & $18.8$ \\
\midrule
\rowcolor{black!5}\multicolumn{9}{@{}l}{\textbf{Circle Packing}} \\
ShinkaEvolve & $6.11$ & -- & $150.0$ & $100.0$ & -- & -- & -- & -- \\
One-Step Alloc & $10.57$ & $0.38$ ($3.6$\%) & $266.8$ & $41.3$ & -- & -- & 41.6 & 17.1 \\
Lineage-UCB & $6.61$ & -- & $190.4$ & $78.8$ & -- & -- & $21.2$ & -- \\
RPM & $121.31$ & $29.64$ ($24.6$\%) & $2250.0$ & $6.7$ & -- & -- & $93.3$ & -- \\
\ours{EvoAlloc (ours)} & $10.85$ & $1.67$ ($16.1$\%) & $243.0$ & $25.5$ & -- & -- & $58.5$ & $16.0$ \\
\bottomrule
\end{tabular*}
\end{table}

\clearpage
\subsection{Additional Ablation Experiments}
\label{app:additional-ablations}

\paragraph{Ablation of Random Allocation.}
\label{app:random-allocation}

To examine whether content-independent filtering can reproduce
EvoAlloc's gains, we evaluate Random Allocation on Circle Packing.
After warm-up, this allocator discards each candidate with probability
$0.5$, independently of its content, and otherwise requests Full evaluation.
We compare five-seed results under a budget of 150 Full evaluations
per run, using the main-experiment EvoAlloc results as the reference.

\begin{figure}[H]
    \centering
    \includegraphics[width=0.80\linewidth]
    {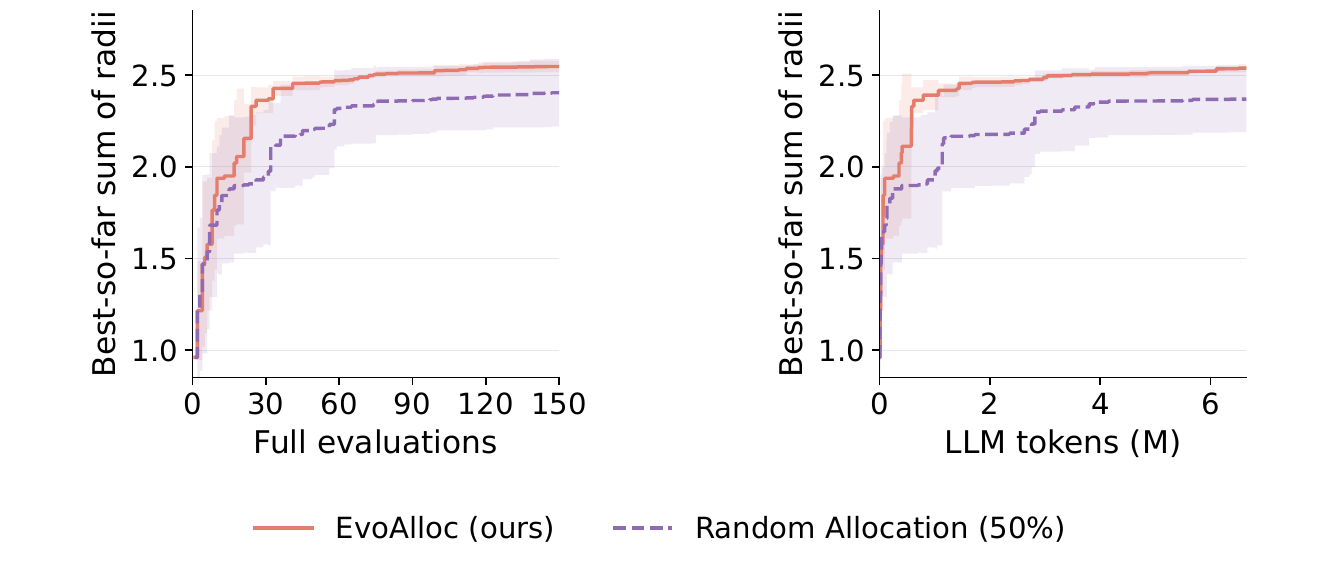}
    \caption{\textbf{Ablation with random allocation on Circle Packing.}
    Random Allocation discards candidates with probability $0.5$
    after warm-up.
    Best-so-far performance is shown against
    Full evaluations (left) and cumulative LLM tokens (right).
    Lines show five-seed means, and shading denotes
    $\pm 1$ standard deviation.
    Token curves use the common observed range across methods and seeds.}
    \label{fig:random-allocation}
\end{figure}

EvoAlloc achieves a final sum of radii of $2.547 \pm 0.030$, compared
with $2.404 \pm 0.184$ for Random Allocation.
Figure~\ref{fig:random-allocation} also shows higher mean best-so-far
performance for EvoAlloc against both Full evaluations and cumulative
LLM tokens.
Randomly withholding evaluations at this fixed probability therefore
does not reproduce EvoAlloc's observed performance.

\paragraph{Ablation of Exploration Feedback.}
\label{app:exploration-feedback}

The \textbf{w/o Exploration} ablation in Section~\ref{sec:controlled_design_main} removes exploratory
evaluations together with the feedback they provide.
To examine the feedback contribution while retaining exploration,
we introduce \textbf{w/o Exploration Feedback}.
This variant keeps the exploration probability at $\rho=0.3$:
explored candidates still receive Full evaluation, enter the search
archive, and can improve the best-so-far score or serve as parents.
However, their outcome cases are withheld from subsequent allocation
decisions, Experience updates, and Strategy revisions.
The ablation thus removes explicit case feedback while preserving
explored candidates' participation in program search.

\begin{figure}[H]
    \centering
    \includegraphics[width=0.80\linewidth]
    {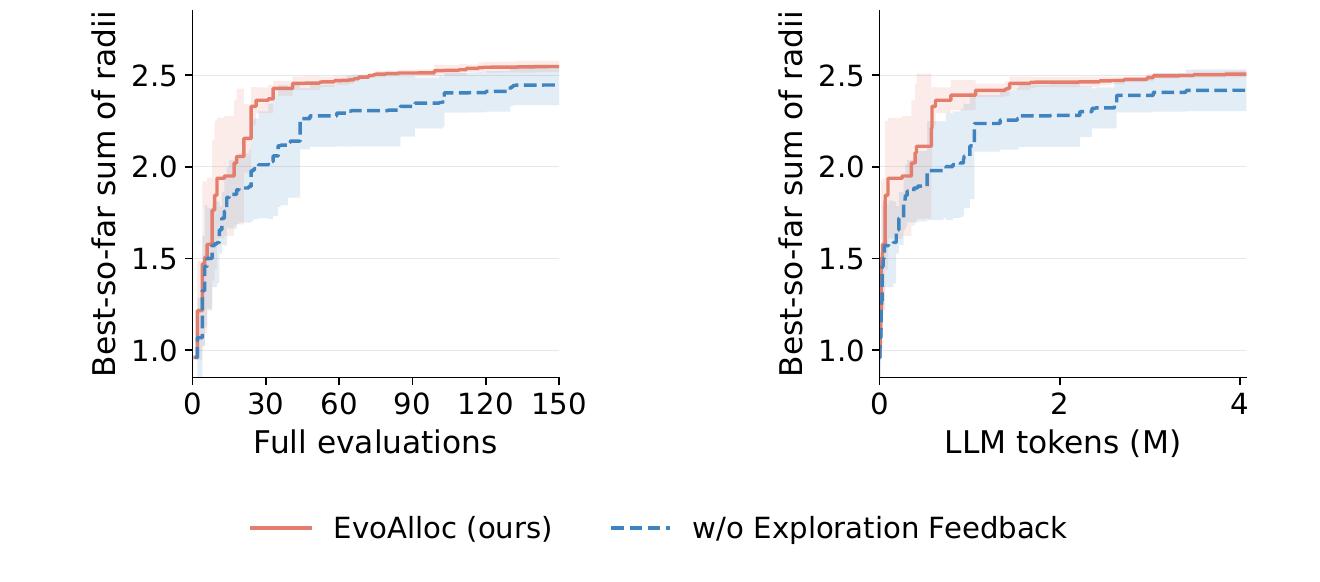}
    \caption{\textbf{Ablation of exploration feedback on Circle Packing.}
    The ablated variant retains exploratory evaluations and archive
    updates but withholds their explicit outcome-case feedback from
    allocation and self-evolution.
    Best-so-far performance is shown against
    Full evaluations (left) and cumulative LLM tokens (right).
    Lines show five-seed means, and shading denotes
    $\pm 1$ standard deviation.
    Token curves use the common observed range across methods and seeds.}
    \label{fig:exploration-feedback}
\end{figure}

We compare five-seed results with the main-experiment EvoAlloc reference,
using budgets of $150$ Full evaluations on Circle Packing and $75$ on
ADAS-AIME.
On Circle Packing, w/o Exploration Feedback achieves
$2.446 \pm 0.110$, compared with $2.547 \pm 0.030$ for EvoAlloc;
the corresponding trajectories are shown in
Figure~\ref{fig:exploration-feedback}.
On ADAS-AIME, the final accuracies are
$(32.22 \pm 4.10)\%$ and $(39.56 \pm 6.80)\%$, respectively.
The lower mean performance without explicit feedback is consistent
with exploration helping not only by recovering candidates, but also
by informing subsequent allocation.

\subsection{Exploration Outcomes across Strategy Stages}
\label{app:exploration-outcomes}

We analyze candidates with Full outcomes obtained through random
counterfactual exploration ($\rho = 0.3$) from five runs per benchmark.
We count only exploration following \textsc{Discard} or \textsc{Stop},
excluding warm-up and shared-validation evaluations.
Candidates are grouped by the Strategy active at decision time:
$S_0$ is the initial Strategy, and subsequent indices denote successive
adopted revisions within each run.
A candidate is \emph{new-best} if its Full score exceeds the
decision-time best-so-far score, and \emph{improved-parent} if it
exceeds its parent's score; the latter includes new-best candidates.

Table~\ref{tab:exploration-outcomes} shows that these recoveries are
not confined to the initial Strategy: all six new-best candidates
on ADAS-AIME occur under $S_2$ or $S_3$, while Circle Packing
recovers 15 under $S_2$.
Exploration thus reveals missed improvements even after multiple
Strategy revisions, exposing outcomes that the allocator's decisions
alone would leave unobserved.
Stage coverage differs across runs, with only one run per benchmark
reaching $S_3$, so the pooled rates should not be interpreted as
a common temporal trend.

\begin{table}[htbp]
\centering
\caption{
\textbf{Outcomes recovered by counterfactual exploration
across Strategy stages.}
Counts are pooled across five seeds within each benchmark.
Percentages use the number of explored candidates in each
row as the denominator, rather than averaging per-seed rates.
Improved-parent includes new-best.
}
\label{tab:exploration-outcomes}
\small
\setlength{\tabcolsep}{4pt}
\renewcommand{\arraystretch}{1.1}
\begin{tabular}{llrrrr}
\toprule
Benchmark
& \shortstack[l]{Strategy\\stage}
& \shortstack[r]{Runs reaching\\the stage}
& \shortstack[r]{Explored\\candidates}
& New-best
& Improved-parent \\
\midrule
\multirow{5}{*}{ADAS-AIME}
& $S_0$ & $5$ & $10$
& $0$ ($0.0$\%) & $3$ ($30.0$\%) \\
& $S_1$ & $5$ & $35$
& $0$ ($0.0$\%) & $4$ ($11.4$\%) \\
& $S_2$ & $4$ & $72$
& $5$ ($6.9$\%) & $18$ ($25.0$\%) \\
& $S_3$ & $1$ & $15$
& $1$ ($6.7$\%) & $8$ ($53.3$\%) \\
\cmidrule(l){2-6}
& \textbf{Total} & -  & $132$
& $6$ ($4.5$\%) & $33$ ($25.0$\%) \\
\midrule
\multirow{5}{*}{Circle Packing}
& $S_0$ & $5$ & $13$
& $2$ ($15.4$\%) & $7$ ($53.8$\%) \\
& $S_1$ & $5$ & $80$
& $5$ ($6.3$\%) & $22$ ($27.5$\%) \\
& $S_2$ & $5$ & $200$
& $15$ ($7.5$\%) & $62$ ($31.0$\%) \\
& $S_3$ & $1$ & $3$
& $0$ ($0.0$\%) & $0$ ($0.0$\%) \\
\cmidrule(l){2-6}
& \textbf{Total} & - & $296$
& $22$ ($7.4$\%) & $91$ ($30.7$\%) \\
\bottomrule
\end{tabular}
\end{table}

\newpage
\subsection{Strategy Validation Details}
\label{app:analysis}

\paragraph{Shared evaluation.}
During validation, the incumbent Strategy $S$ and the challenger $\widetilde{S}$
are applied to the same online proposals with the same archive, Experiences, and
decision-time information. EvoAlloc executes the union of their requested evaluations
once, charges them to the search budget, and shares the resulting evidence with both
Strategies. A candidate requested for Full evaluation only by the challenger is therefore
still evaluated, so validation is not restricted to outcomes selected by the incumbent.
Outcomes of candidates that neither Strategy sends to Full evaluation are revealed only
through counterfactual exploration (Section~\ref{sec:counterfactual_correction}), which also supplies evidence for
skipped candidates outside validation windows.

\paragraph{Consequence-based comparison.}
Let $I$ denote the proposals in a validation window
on which the two Strategies select different allocation paths and the Full outcome is
known, whether requested by either Strategy or revealed by exploration; a window
closes once $|I|=2$. For $T\in\{S,\widetilde{S}\}$, let $m_{\mathrm{best}}(T)$ count proposals in $I$ whose Full
score exceeds the best score in the shared archive at decision time but whose allocation
path under $T$ would leave that outcome hidden. Let $n_F(T)$ and $n_P(T)$ count the Full
and Partial evaluations requested under $T$ on these proposals, rather than those
executed for the union. The challenger is adopted if
\[
Z(T)=\bigl(m_{\mathrm{best}}(T),\,n_F(T),\,n_P(T)\bigr)
\]
is lexicographically smaller for $\widetilde{S}$ than for $S$; a complete tie retains the incumbent, and
the last component is omitted when no Partial evaluator is available. This establishes a
preference over realized allocation consequences within a finite window, not a
guarantee of improvement in $\mathbb{E}[Q^\star_{\mathcal{A}}(B)]$; continued search supplies evidence for
subsequent Strategy revisions.

\paragraph{Observed validation outcomes.}
Table~\ref{tab:validation-stats} summarizes how often proposed challengers were
adopted across both benchmarks, together with the distribution of adoption
criteria. Each validation window compares a challenger with the incumbent on
shared proposals; completed windows are those that reached a decision to adopt
the challenger or retain the incumbent.
The reported adoptions were driven by fewer Full or Partial evaluations, while
matching the incumbent's missed-new-best count within the corresponding
validation windows. Validation thus primarily favored more selective use of
evaluation resources, refining how much evaluation to allocate to each candidate.

\begin{table}[H]
\centering
\caption{Strategy validation outcomes aggregated over five seeds per benchmark.
The final three columns count adoptions decided by fewer missed new-best
candidates ($m_{\mathrm{best}}$), fewer Full evaluations ($n_F$), or fewer
Partial evaluations ($n_P$).}
\label{tab:validation-stats}
\small
\setlength{\tabcolsep}{3pt}
\resizebox{\linewidth}{!}{%
\begin{tabular}{@{}lccc*{3}{>{\centering\arraybackslash}p{1.05cm}}@{}}
\toprule
& & & & \multicolumn{3}{c}{Adoption criterion} \\
\cmidrule(lr){5-7}
Benchmark
& Proposed challengers
& Completed windows
& Adopted challengers
& $m_{\mathrm{best}}$
& $n_F$
& $n_P$ \\
\midrule
ADAS-AIME & $23$ & $20$ & $10$ & $0$ & $7$ & $3$ \\
Circle Packing & $44$ & $42$ & $11$ & $0$ & $11$ & $0$ \\
\bottomrule
\end{tabular}}
\end{table}

%% file: sections/algorithm.tex
\begingroup
\renewcommand{\eqref}[1]{\textup{(\ref{#1})}}
\newcommand{\fn}[1]{\textsc{#1}}
\newcommand{\act}[1]{\textsc{#1}}
\newcommand{\cact}[1]{\mbox{\small\texttt{#1}}}
\newcommand{\kw}[1]{\textbf{#1}}
\definecolor{evocommentgreen}{RGB}{66,128,89}
\newcommand{\cmt}[1]{%
  \hfill{\small\color{evocommentgreen}$\triangleright$\ #1}}
\newcommand{\tightcmt}[1]{%
  \hfill{\footnotesize\color{evocommentgreen}$\triangleright$\ #1}}
\newcommand{\bcmt}[1]{%
  \hfill{\small\color{evocommentgreen}$\triangleright$\ #1}}
\newcommand{\ai}{\hspace*{1.5em}}
\newcommand{\aii}{\hspace*{3.0em}}
\newcommand{\aiii}{\hspace*{4.5em}}
\definecolor{evoyellow}{RGB}{250,247,214}
\newcommand{\evorow}{\rowcolor{evoyellow}}

Algorithm~\ref{alg:evoalloc} highlights EvoAlloc-specific operations.
Parent selection, proposal generation, archive updates, and benchmark evaluation
follow the underlying search procedure.

\begin{center}
\refstepcounter{algocf}\label{alg:evoalloc}
\begin{minipage}{\linewidth}
\hrule height 1pt
\vspace{3pt}

{\bfseries Algorithm~\thealgocf:}\;
Program evolution with EvoAlloc\\[-2pt]

\hrule height 0.5pt
\vspace{3pt}

\normalsize
\noindent\kw{Input:}
\textit{initial programs};
\textit{example cases};
\textit{Full-evaluation budget} $B$;
\textit{exploration rate} $\rho$;
\textit{update intervals} $k_E, k_S$.
\par\vspace{2pt}

\noindent\kw{Output:}
\textit{best program} $x^*$
\par\vspace{4pt}

\hrule height 0.4pt
\vspace{3pt}

\begin{tabularx}{\linewidth}{@{}r@{\quad}>{\raggedright\arraybackslash}X@{}}
1 &
$\mathcal{P}\leftarrow\fn{InitArchive}(\text{initial programs})$;\quad
$H\leftarrow\emptyset$;\quad $n\leftarrow0$
\cmt{archive and interaction history} \\
\evorow
2 & $S\leftarrow S_0\leftarrow\fn{InitStrategy}(\text{example cases})$;\quad
$E\leftarrow\emptyset$;\quad $\tilde{S}\leftarrow\emptyset$ \\[1pt]
3 & \kw{while} $n<B$ \kw{do} \\
4 & \ai $x_p\leftarrow\fn{SelectParent}(\mathcal{P})$;\quad
$x\leftarrow\fn{GenerateProposal}(x_p,\mathcal{P})$ \\
\evorow
5 & \ai $c\leftarrow\fn{BuildContext}(x_p,\mathcal{P},H,n,B)$
\cmt{decision-time context} \\
\evorow
6 & \ai $\tau\leftarrow\fn{Allocate}(x,c;S,E)$
\cmt{executes and caches any Partial evaluation} \\
\evorow
7 & \ai $\tilde{\tau}\leftarrow\fn{Allocate}(x,c;\tilde{S},E)$ \kw{if} $\tilde{S}\neq\emptyset$ \kw{else} $\emptyset$\newline
\makebox[\linewidth][r]{{\small\color{evocommentgreen}$\triangleright$\ shadow allocation; reuses cached Partial evidence}} \\
\evorow
8 & \ai \kw{if} $\tau$ ends in Full \kw{then} \\
\evorow
9 & \aii $y^{\mathrm{obs}}\leftarrow\fn{FullEval}(x)$ \\
\evorow
10 & \ai \kw{else if} \kw{not} $\fn{WarmupComplete}(H)$ \kw{then} \\
\evorow
11 & \aii $y^{\mathrm{obs}}\leftarrow\fn{FullEval}(x)$ \\
\evorow
12 & \ai \kw{else if} $\tilde{\tau}$ ends in Full \kw{then} \\
\evorow
13 & \aii $y^{\mathrm{obs}}\leftarrow\fn{FullEval}(x)$
\cmt{shared validation (Sec.~\ref{sec:allocation_state})} \\
\evorow
14 & \ai \kw{else} \\
\evorow
15 & \aii sample $z\sim\mathrm{Bernoulli}(\rho)$ \\
\evorow
16 & \aii \kw{if} $z=1$ \kw{then} $y^{\mathrm{obs}}\leftarrow\fn{FullEval}(x)$ \\
\evorow
17 & \aii \kw{else} $y^{\mathrm{obs}}\leftarrow\emptyset$ \\
\evorow
18 & \ai $H\leftarrow H\cup\{(x,c,\tau,\widetilde{\tau},y^{\mathrm{obs}})\}$ \\
19 & \ai \kw{if} $y^{\mathrm{obs}}\neq\emptyset$ \kw{then}
$n\leftarrow n+1$;\quad
$\mathcal{P}\leftarrow\fn{UpdateArchive}(\mathcal{P},x,y^{\mathrm{obs}})$ \\
\evorow
20 & \ai \kw{if} $k_E$ new Full outcomes since the last Experience update \kw{then}\newline
\aii $E\leftarrow\fn{UpdateExperiences}(E,H)$ \\
\evorow
21 & \ai \kw{if} $\tilde{S}\neq\emptyset$ \kw{and}
$\fn{ValidationComplete}(H)$ \kw{then} \\
\evorow
22 & \aii $S\leftarrow\fn{ValidateStrategy}(S,\tilde{S},H)$;\quad
$\tilde{S}\leftarrow\emptyset$
\cmt{adopt or retain} \\
\evorow
23 & \ai \kw{else if} $\tilde{S}=\emptyset$ \kw{and} $\fn{WarmupComplete}(H)$ \kw{and}\newline
\aii $k_S$ new Full outcomes since the last reflection \kw{then} \\
\evorow
24 & \aii $\tilde{S}\leftarrow\fn{ReflectStrategy}(S,E,H)$
\cmt{one challenger at a time} \\
25 & \kw{return} $\fn{Best}(\mathcal{P})$ \\
\end{tabularx}

\vspace{3pt}
\hrule height 1pt
\vspace{4pt}

{\small
Here, $\mathcal{P}$ is the search archive and $H$ the interaction history
(Section~\ref{sec:formulation:strategy-evolution}), retaining original allocation episodes even when outcomes
are later revealed.
\textsc{Allocate} follows the allocation procedure in Section~\ref{sec:allocation-strategy}.
During validation, the union of requested evaluations is executed once
and the resulting evidence is shared.
All Full evaluations, including warm-up, validation, and exploration,
count toward $B$.
}
\end{minipage}
\end{center}
\endgroup

%% file: sections/benchmark_protocols.tex
\subsection{ADAS-AIME}
\label{app:adas_program}

ADAS-AIME evolves the Python scaffold around a fixed mathematical-reasoning
solver on the 30 AIME~2024 problems. A candidate is a reusable
\texttt{Agent}: it may query the solver, vary sampling temperature, combine
responses, filter or vote over candidate answers, and extract the final integer
answer returned to the benchmark harness. The solver weights and benchmark
questions are fixed; the evolved object is the reasoning scaffold.

Following the ShinkaEvolve evaluation protocol, each candidate is evaluated over all 30 AIME 2024
problems in three independent runs, and the resulting accuracies are averaged. These three repetitions
together count as one Full evaluation in the search budget. For allocation decisions, we use a Partial evaluation on the first 8 of 30 problems using a single evaluation seed.
After observing the Partial result, the allocator chooses \texttt{CONTINUE\_TO\_FULL} or \texttt{STOP}.
The Partial result is used only to decide whether the current proposal deserves the remaining full-evaluation
cost; final performance is always measured by the full evaluator.
Search and evaluation use the same fixed 30-problem suite.

The example below illustrates a reasoning harness that samples eight solutions
at three temperatures, extracts candidate answers, and selects the most frequent
answer by majority voting.

\noindent\hspace*{0.03\linewidth}{\ttfamily\bfseries\footnotesize
Example evolved program: ADAS-AIME}\par\vspace{2pt}
\begin{codebox}[label={lst:adas_best}, width=\linewidth]
import re
from collections import Counter
from typing import Callable, Optional

class Agent:
    def __init__(
        self,
        query_llm: Callable,
        temperature=0.0,
    ):
        self.query_llm = query_llm
        self.max_calls = 10
    def _extract_answer(self, response: str) -> Optional[int]:
        match = re.search(r'\\boxed\{(\d+)\}', response)
        if match:
            return int(match.group(1))
        lines = response.strip().split('\n')
        for line in reversed(lines):
            line = line.strip()
            if line.isdigit() and 0 <= int(line) <= 999:
                return int(line)
        return None
    def forward(self, problem: str) -> tuple[str, float]:
        system_prompt = (
            "You are an expert mathematician. Solve the following problem step-by-step. "
            "Think carefully and verify your steps. "
            "On the final line output only the digits of the answer (0-999). "
            "Provide your final answer enclosed in a LaTeX \\boxed{...} command."
        )
        temperatures = [0.8, 0.8, 0.8, 0.5, 0.5, 0.2, 0.2, 0.2]
        candidates = []
        total_cost = 0.0
        for t in temperatures:
            if len(candidates) >= self.max_calls:
                break
            task_prompt = f"{problem}\n\n"
            response, cost = self.query_llm(
                prompt=task_prompt,
                system=system_prompt,
                temperature=t,
            )
            total_cost += cost
            ans = self._extract_answer(response)
            if ans is not None:
                candidates.append((ans, response))
        if not candidates:
            return "", total_cost
        ans_counts = Counter([c[0] for c in candidates])
        best_ans = ans_counts.most_common(1)[0][0]
        for ans, resp in candidates:
            if ans == best_ans:
                return resp, total_cost
        return candidates[0][1], total_cost
\end{codebox}

\subsection{Circle Packing}
\label{app:circle_program}

Circle Packing evolves compact Python programs that construct feasible
placements of 26 circles in the unit square. The evaluator checks boundary and
non-overlap constraints, computes the induced radii, and returns the packing
objective. Unlike ADAS-AIME, the evaluator is deterministic and non-LLM; the
artifact is a constructive geometric algorithm rather than a language-model
scaffold.

We do not use Partial evaluation in this setting. The allocator therefore
chooses directly between \texttt{FULL\_EVAL} and \texttt{DISCARD}.
This setting isolates adaptive full-evaluation allocation on a deterministic algorithm-design task.

\noindent\hspace*{0.03\linewidth}{\ttfamily\bfseries\footnotesize
Example evolved program (excerpt): Circle Packing}\par\vspace{2pt}
\begin{codebox}[label={lst:circle_best}, width=\linewidth]
def construct_packing():
    n = 26
    rng = np.random.default_rng(seed=12345)
    initializations = [
        _init_hexagonal(n, rng),
        _init_concentric(n, rng),
        _init_corner_focused(n, rng),
        _init_voronoi(n, rng),
        _init_strategic(n, rng),
    ]
    best_centers = None
    best_radii = None
    best_score = -1.0
    for init_c in initializations:
        best_c, best_r = anneal_centers(init_c, rng, max_iter=35000)
        best_r = radius_expansion(best_c, best_r, max_iters=10)
        best_c, best_r = gradient_refine(best_c, best_r, iterations=14, step_size=0.022, momentum=0.78)
        best_r = _lp_max_radii(best_c)
        score = best_r.sum()
        if score > best_score:
            best_score = score
            best_centers = best_c.copy()
            best_radii = best_r.copy()
    best_radii = np.maximum(best_radii, 0.0)
    return best_centers, best_radii
\end{codebox}

%% file: sections/presubmission_checklist.tex
\section{Experimental Details and Additional Results}
\label{app:presubmission_checklist}
\label{app:detailed_results}

This appendix describes the baseline and ablation configurations and reports detailed performance, resource costs, and allocation breakdowns. Implementation settings and benchmark protocols are given in Appendix~\ref{app:impl}.

\subsection{Baselines and Comparison Configuration}
\label{app:baselines_ablations}

All methods run within the same ShinkaEvolve framework, using its default configuration and benchmark evaluation protocols unless stated otherwise in Appendix~\ref{app:impl}. They differ only in how evaluation resources are allocated after a candidate is generated. Following the framework's default behavior, a candidate that does not ultimately receive Full evaluation is discarded and cannot be selected as a parent. Allocation therefore determines which candidates enter the archive, and different methods may consequently induce different proposal streams over the course of a run; we treat this interaction as part of the closed-loop allocation problem. While the default configuration uses a single seed, we run each method with five seeds and report averages for more reliable comparisons.

\paragraph{ShinkaEvolve.}
ShinkaEvolve \citep{lange2025shinkaevolve} is the program-evolution framework underlying all methods and serves as the full-evaluation reference, in which every generated candidate receives Full evaluation. We retain its main search mechanisms, including an ensemble of proposer LLMs sampled at multiple temperatures, a meta model that periodically reflects on evaluated programs to guide subsequent proposals, and novelty filtering that combines embedding similarity with an LLM novelty judge to reject near-duplicate candidates. The models used for each role are listed in Table~\ref{tab:appendix_impl_summary}.
We rerun ShinkaEvolve with the same locally served open-weight search
models used by EvoAlloc and the other baselines (Table~\ref{tab:appendix_impl_summary}).
This differs from the proprietary model ensemble used in the original
study, so the reported scores should be interpreted within our shared
model configuration rather than as a reproduction of its published results.
Our comparisons assess evaluation allocation under a common search
backbone and matched Full-evaluation budgets.

\paragraph{One-Step LLM Allocator.}
This baseline isolates the effect of EvoAlloc's persistent allocation state while matching its allocator, action space, counterfactual exploration, and decision context.
Each decision is conditioned on the same search information available to EvoAlloc, including recent cases, retrieved similar cases, and the current search state.
The baseline differs only in maintaining neither an explicit Strategy nor distilled Experiences; instead, it maps the available context directly to an allocation action at each step.
This comparison evaluates the benefit of explicitly consolidating search history into reusable Experiences and an evolving Strategy, beyond directly conditioning an LLM allocator on the search history.

\paragraph{Lineage-UCB.}
Lineage-UCB represents classical adaptive allocation based on a fixed statistical rule, in the spirit of bandit-based methods. It uses a binary action space of Full evaluation or discard on both benchmarks. For each candidate, it estimates the improvement tendency of its local lineage using a Beta posterior initialized as $\mathrm{Beta}(1,1)$ and updated from previously observed outcomes, with weight $1$ for direct relatives and $0.5$ for second-level relatives. A candidate receives Full evaluation if the posterior mean plus one standard deviation is at least $0.5$, and is otherwise discarded. The allocator requires no LLM calls. This comparison isolates the benefit of learned, structured allocation guidance beyond online adaptation under a fixed statistical rule.

\paragraph{RPM.}
We include RPM \citep{foster2026rpm} as a recent LLM-based proposal-selection
baseline that addresses the closely related problem of deciding which
candidates merit computationally expensive evaluation. We adapt its inference-time selector to our setting. At each selection step, 15 candidates are generated from the same parent and compared through a seeded, randomly ordered knockout tournament, with the winner receiving Full evaluation and all others discarded. The batch size of 15 follows the end-to-end configuration in the original RPM study. Each pairwise comparison is conditioned on the two candidates' plans and code, together with up to 10 previously evaluated programs retrieved from the search history. We use Qwen3-Coder-Next as the judge with temperature 0 and a maximum of 4{,}096 output tokens. Unlike EvoAlloc, RPM performs batch-level selection among competing candidates and follows a fixed selection rule throughout search. LLM tokens consumed by all generated candidates, including those not selected for evaluation, are included in the total cost.

\paragraph{Ablations.}
Section~\ref{sec:controlled_design_main} compares EvoAlloc with variants using one-step allocation, fixed allocation guidance, and no counterfactual exploration. \textbf{One-Step Alloc} corresponds to the One-Step LLM Allocator described above, removing both the persistent Strategy and distilled Experiences. \textbf{w/o Evolution} uses fixed custom allocation guidance: a medium-selectivity Strategy and seeded Experiences distilled from baseline run examples, with the Strategy manually refined into reasonable allocation criteria. This guidance remains fixed throughout search. \textbf{w/o Exploration} sets \mbox{$\rho=0$}, preventing discarded or stopped candidates from receiving exploratory Full evaluation and thus leaving their Full outcomes unobserved.

%% file: sections/artifacts_prompts.tex
\section{Prompts and Context}
\label{app:artifacts_prompts}
\label{app:prompts}

This appendix details the inputs, prompts, and representative outputs of the two LLM calls in EvoAlloc: resource allocation for each candidate (Section~\ref{sec:allocation_strategy}), and self-evolution, which updates Experiences and proposes Strategy revisions (Section~\ref{sec:strategy-evolution}).

\subsection{Input Context}
\label{app:memory_context}

Each LLM call in EvoAlloc is conditioned on a structured context assembled from the ongoing search, summarized in Table~\ref{tab:input_context}. Resource allocation and self-evolution both receive the benchmark setup, the current allocation guidance, recent cases, and a summary of search progress. Resource allocation additionally receives the current candidate, its search context, and historical cases retrieved by similarity; together with search progress, these form the decision-time context $c_t$ of Section~\ref{sec:stateful_allocation}. Since allocation decisions are made before the candidate is fully evaluated, its own Full outcome is never part of this context.

\begin{table}[h]
\centering
\small
\caption{\textbf{Input context for resource allocation and self-evolution.}}
\label{tab:input_context}
\label{tab:evocritic_input_context}
\begin{tabularx}{\linewidth}{@{}l X@{}}
\toprule
\textbf{Context} & \textbf{Information} \\
\midrule
\multicolumn{2}{@{}l}{\textit{Shared by resource allocation and self-evolution}} \\
\midrule
Benchmark & Evaluation protocol and available actions \\
Allocation guidance & Current Strategy and Experiences \\
Recent cases & Recently observed cases, as case cards (Appendix~\ref{app:case_cards}) \\
Search progress & Budget usage and recent decision counts \\
\midrule
\multicolumn{2}{@{}l}{\textit{Additional context for resource allocation}} \\
\midrule
Candidate & Candidate description and code changes relative to its parent \\
Search context & Lineage scores, the parent's gap to the current best, and sibling outcomes \\
Retrieved cases & Similar historical cases, with their observed outcomes \\
Partial evidence & Partial-evaluation result, for the continuation decision only \\
\bottomrule
\end{tabularx}
\end{table}

\subsection{Resource allocation prompt}
\label{app:allocator_prompt}

The prompt below guides resource allocation using the current Strategy,
relevant Experiences, and available search context. Representative structured
outputs illustrate the initial allocation and post-Partial continuation decisions.

\begin{promptlisting}[before skip=4pt,top=6pt,bottom=6pt]{Resource allocation prompt}
You are the resource allocator of EvoAlloc, a resource-allocation system in a self-improving program search loop. EvoAlloc aims to improve sample efficiency over the full run by allocating evaluations where they have the greatest search value.

Through self-evolution, EvoAlloc learns reusable search experiences from observed outcomes and maintains an allocation strategy for future resource allocation. Use this strategy as the primary guidance, together with relevant experiences and search context, to choose among:

- FULL_EVAL: run the full benchmark evaluation and observe its result.
- PARTIAL_EVAL: obtain limited evidence, then choose CONTINUE_TO_FULL or STOP.
- DISCARD: skip full evaluation and leave its result hidden.

Judge the proposal's marginal search value against preserving the same resources for future search. Consider both its current evidence and whether evaluating it would preserve a branch with meaningful downstream search value.

Use the allocation strategy as the run-level default, and adjust the decision when current grounded experience is materially relevant.

Apply only guidance that is usable at decision time; ignore any condition that requires the current proposal's unknown Full outcome.

- Use PARTIAL_EVAL only when at least two plausible partial-evaluation observations would support different final actions. If likely observations would all support the same final action, take it now. Partial evaluation is an information-acquisition action, not a safe compromise or an automatic step before Full evaluation.

- Interpret partial evaluation according to what it reveals: successful execution alone does not justify CONTINUE_TO_FULL, and low partial performance alone does not justify STOP. A reproducible candidate-side execution failure is strong negative evidence because Full evaluation runs the same candidate unchanged.
\end{promptlisting}

\paragraph{Post-Partial continuation.}
\label{app:smoke_prompt}

When Partial evaluation is requested, its verified observation is appended to the original
allocation context and the allocator makes a second decision using the same
proposal, Strategy, Experiences, and search context.

\begin{smokepromptlisting}{Post-Partial continuation prompt}
The partial evaluator returned the following verified observation:
{partial_observation}

Complete the same allocation episode using the original proposal, allocation strategy, experiences, and search context.

Return only one valid JSON object:
{continuation_schema}
\end{smokepromptlisting}

The outputs use structured JSON. Ellipses in the abbreviated examples denote
generated text.

\noindent
\begin{minipage}[t]{0.49\linewidth}
\vspace{0pt}
\begin{schemalisting}[breakable=false,fonttitle=\footnotesize\bfseries,top=5pt,bottom=5pt,before skip=4pt,after skip=4pt]{Resource allocation: initial decision}
{
  "action": "PARTIAL_EVAL",
  "matched_experience_ids": ["e1", "e3"],
  "reason": "...",
  "partial_request": {"purpose": "..."}
}
\end{schemalisting}
\end{minipage}\hfill
\begin{minipage}[t]{0.49\linewidth}
\vspace{0pt}
\begin{schemalisting}[breakable=false,fonttitle=\footnotesize\bfseries,top=5pt,bottom=5pt,before skip=4pt,after skip=4pt]{Resource allocation: post-Partial}
{
  "action": "CONTINUE_TO_FULL",
  "reason": "...",
  "used_experience_ids": ["e1"]
}
\end{schemalisting}
\end{minipage}

\subsection{Self-Evolution Prompts}
\label{app:evolution_prompt}

The prompts below specify separate updates to Experiences and the Strategy.
Experience updates maintain case-grounded evidence, while Strategy updates
propose challengers for shadow validation.

\paragraph{Experience operations.}
\label{app:experience-ops}
\texttt{ADD} introduces a new Experience with initial status \texttt{tentative}.
\texttt{UPDATE} attaches supporting or contradicting evidence to an existing Experience and may change its status, but cannot rewrite its content, so accumulated evidence always refers to a fixed statement.
\texttt{DELETE} removes an Experience.
Rewriting an Experience therefore requires deleting it and adding a new one, and no operation is performed when evidence is insufficient.
Each Experience has one of three statuses: \texttt{tentative} (limited or mixed evidence), \texttt{active} (consistent evidence), or \texttt{stale} (retained as a historical record but no longer guiding current decisions).
Unlike \texttt{DELETE}, marking an Experience \texttt{stale} keeps it in memory.

\begin{criticpromptlisting}[before skip=5pt,after skip=5pt,top=6pt,bottom=6pt]{Experience update prompt}
Update only the evolving search experiences.

Experiences are descriptive. Each contains one recurring pattern and one qualitative outcome tendency, separated into applicability visible at decision time and observed tendency. It must remain identifiable and useful across future proposals, while retaining enough mechanism-level detail to distinguish the pattern. Experiences do not prescribe allocation actions.

Full outcomes, including those revealed by exploration, are grounded evidence. Explored outcomes provide grounded evidence for allocation decisions whose Full outcomes would otherwise remain hidden. Partial evaluation provides limited evidence; unexplored DISCARD and STOP cases have unknown Full outcomes.
\end{criticpromptlisting}

\begin{criticpromptlisting}[before skip=5pt,after skip=5pt,top=6pt,bottom=6pt]{Strategy update prompt}
Update only the allocation strategy.

The strategy is prescriptive. It is the allocator's run-level posture for balancing the available allocation actions as evidence and budget change, rather than encoding a decision tree.

Reconsider the strategy using recent outcomes, including those revealed by exploration, together with accumulated experiences and broader run context, then ALWAYS propose the allocation strategy you would use right now as a challenger.
\end{criticpromptlisting}

\noindent
\begin{minipage}[t]{0.49\linewidth}
\vspace{0pt}
\begin{schemalisting}[breakable=false,fonttitle=\footnotesize\bfseries,top=5pt,bottom=5pt,before skip=4pt,after skip=4pt]{Self-evolution: Experience update}
{
  "experience_operations": [{
      "operation": "UPDATE",
      "experience_id": "e1",
      "status": "active",
      "support_case_ids":
        ["Case-5", "Case-10"],
      "counter_case_ids": [],
      "rationale": "..."
  }]
}
\end{schemalisting}
\end{minipage}\hfill
\begin{minipage}[t]{0.49\linewidth}
\vspace{0pt}
\begin{schemalisting}[breakable=false,fonttitle=\footnotesize\bfseries,top=5pt,bottom=5pt,before skip=4pt,after skip=4pt]{Self-evolution: Strategy update}
{
  "allocation_strategy_update": {
    "operation": "REVISE",
    "selectivity": "medium",
    "allocation_strategy": "...",
    "rationale": "...",
    "evidence_experience_ids": ["e1"]
  }
}
\end{schemalisting}
\end{minipage}

\subsection{Case Card Examples}
\label{app:case_cards}

Each observed candidate is summarized as a case card, which records the proposal, its search context, the allocation decision, and the observed outcomes. Case cards provide the recent and retrieved cases in the input context of resource allocation and self-evolution (Table~\ref{tab:input_context}). The examples below show two ADAS-AIME case cards: Partial evaluation followed by Full evaluation, and Partial evaluation followed by \texttt{STOP} and later counterfactual exploration.

\begin{center}
\begin{minipage}{0.82\linewidth}
\begin{casecardbox}[label={lst:case_card_adas}]{ADAS-AIME: Partial evaluation followed by Full evaluation}
Case-11

Proposal:
- Name: self_consistency_ensemble_verification

Search context:
- Current best: 32.222
- Lineage scores: 21.111
- Parent position: below current best
- Lineage trend: insufficient observed history
- Earlier siblings: 1 observed; scores 32.222
- Sibling outcomes: 1/1 improved over parent; 1/1 reached current best
- Decision-time experience matches: e1

Observed allocation path and outcomes:
- Allocation path: PARTIAL_EVAL -> CONTINUE_TO_FULL
- Partial: 3/8 passed; execution completed.
- Full eval: 40.000, valid new-best
\end{casecardbox}
\end{minipage}
\end{center}

\newpage
\begin{center}
\begin{minipage}{0.82\linewidth}
\begin{casecardbox}[label={lst:case_card_adas_smoke}]{ADAS-AIME: Partial--STOP followed by counterfactual exploration}
Case-67

Proposal:
- Name: rigorous_verify_synthesize

Search context:
- Current best: 33.333
- Lineage scores: 25.556 -> 28.889 -> 27.778
- Parent position: below current best
- Lineage trend: mixed
- Earlier siblings: 1 observed; scores 27.778
- Sibling outcomes: 0/1 improved over parent; 0/1 reached current best
- Decision-time experience matches: e2

Observed allocation path and outcomes:
- Allocation path: PARTIAL_EVAL -> STOP -> EXPLORE_EVAL
- Partial: 3/8 passed; execution completed.
- Full eval: 28.889, valid improved parent
\end{casecardbox}
\end{minipage}
\end{center}

%% file: refs.bib
@inproceedings{cramer1985representation,
  author    = {Cramer, Nichael Lynn},
  title     = {A Representation for the Adaptive Generation of Simple Sequential Programs},
  booktitle = {Proceedings of the First International Conference on Genetic Algorithms and Their Applications},
  pages     = {183--187},
  year      = {1985}
}

@techreport{dickmanns1987genetische,
  author      = {Dickmanns, Dirk and Schmidhuber, J{\"u}rgen and Winklhofer, Andreas},
  title       = {{Der genetische Algorithmus: Eine Implementierung in Prolog}},
  institution = {Institut f{\"u}r Informatik, Lehrstuhl Prof. Radig, Technische Universit{\"a}t M{\"u}nchen},
  type        = {Fortgeschrittenenpraktikum},
  year        = {1987}
}

@mastersthesis{schmidhuber1987,
  author = {J{\"u}rgen Schmidhuber},
  title  = {Evolutionary Principles in Self-Referential Learning, or on Learning How to Learn: The Meta-Meta-... Hook},
  type   = {Diploma thesis},
  school = {Technische Universit{\"a}t M{\"u}nchen},
  year   = {1987}
}

@incollection{godelmachine,
  author    = {J{\"u}rgen Schmidhuber},
  title     = {{G{\"o}del Machines}: Fully Self-Referential Optimal Universal Self-Improvers},
  booktitle = {Artificial General Intelligence},
  publisher = {Springer},
  pages     = {199--226},
  year      = {2007},
  note      = {Also arXiv:cs/0309048 (2003)}
}

@inproceedings{dgm,
  author  = {Jenny Zhang and Shengran Hu and Cong Lu and Robert Tjarko Lange and Jeff Clune},
  title   = {{Darwin {G}{\"o}del Machine}: Open-Ended Evolution of Self-Improving Agents},
  booktitle = {International Conference on Learning Representations (ICLR)},
  year    = {2026}
}

@inproceedings{hgm,
  author  = {Wenyi Wang and Piotr Pi{\k{e}}kos and {Li Nanbo} and Firas Laakom and Yimeng Chen and Mateusz Ostaszewski and Mingchen Zhuge and J{\"u}rgen Schmidhuber},
  title   = {{Huxley-{G}{\"o}del Machine}: Human-Level Coding Agent Development by an Approximation of the Optimal Self-Improving Machine},
  booktitle = {International Conference on Learning Representations (ICLR)},
  year    = {2026}
}

@article{sia,
  author  = {Maxime Robeyns and Martin Szummer and Laurence Aitchison},
  title   = {A Self-Improving Coding Agent},
  journal = {arXiv preprint arXiv:2504.15228},
  year    = {2025}
}

@article{funsearch,
  author  = {Bernardino Romera-Paredes and Mohammadamin Barekatain and Alexander Novikov and Matej Balog and M. Pawan Kumar and Emilien Dupont and Francisco J. R. Ruiz and Jordan S. Ellenberg and Pengming Wang and Omar Fawzi and Pushmeet Kohli and Alhussein Fawzi},
  title   = {Mathematical Discoveries from Program Search with Large Language Models},
  journal = {Nature},
  volume  = {625},
  pages   = {468--475},
  year    = {2024}
}

@article{shinkaevolve,
  author  = {Robert Tjarko Lange and Yuki Imajuku and Edoardo Cetin},
  title   = {{ShinkaEvolve}: Towards Open-Ended and Sample-Efficient Program Evolution},
  journal = {arXiv preprint arXiv:2509.19349},
  year    = {2025}
}

@article{jin2011saea,
  author  = {Yaochu Jin},
  title   = {Surrogate-Assisted Evolutionary Computation: Recent Advances and Future Challenges},
  journal = {Swarm and Evolutionary Computation},
  volume  = {1},
  number  = {2},
  pages   = {61--70},
  year    = {2011}
}

@inproceedings{snoek2012bo,
  author    = {Jasper Snoek and Hugo Larochelle and Ryan P. Adams},
  title     = {Practical {Bayesian} Optimization of Machine Learning Algorithms},
  booktitle = {Advances in Neural Information Processing Systems},
  volume    = {25},
  pages     = {2951--2959},
  year      = {2012}
}

@article{li2018hyperband,
  author  = {Lisha Li and Kevin Jamieson and Giulia DeSalvo and Afshin Rostamizadeh and Ameet Talwalkar},
  title   = {Hyperband: A Novel Bandit-Based Approach to Hyperparameter Optimization},
  journal = {Journal of Machine Learning Research},
  volume  = {18},
  number  = {185},
  pages   = {1--52},
  year    = {2018}
}

@inproceedings{li2020asha,
  author    = {Liam Li and Kevin Jamieson and Afshin Rostamizadeh and Ekaterina Gonina and Jonathan Ben-tzur and Moritz Hardt and Benjamin Recht and Ameet Talwalkar},
  title     = {A System for Massively Parallel Hyperparameter Tuning},
  booktitle = {Proceedings of Machine Learning and Systems},
  volume    = {2},
  pages     = {230--246},
  year      = {2020}
}

@article{novikov2025alphaevolve,
  title = {{AlphaEvolve}: A Coding Agent for Scientific and Algorithmic Discovery},
  author = {Novikov, Alexander and V{\~{u}}, Ng{\^a}n and Eisenberger, Marvin and Dupont, Emilien and Huang, Po-Sen and Wagner, Adam Zsolt and Shirobokov, Sergey and Kozlovskii, Borislav and Ruiz, Francisco J. R. and Mehrabian, Abbas and Kumar, M. Pawan and See, Abigail and Chaudhuri, Swarat and Holland, George and Davies, Alex and Nowozin, Sebastian and Kohli, Pushmeet and Balog, Matej},
  journal = {arXiv preprint arXiv:2506.13131}, year = {2025}
}

@inproceedings{lange2025shinkaevolve,
  title = {{ShinkaEvolve}: Towards Open-Ended and Sample-Efficient Program Evolution},
  author = {Lange, Robert Tjarko and Imajuku, Yuki and Cetin, Edoardo},
  booktitle = {International Conference on Learning Representations (ICLR)},
  year = {2026}
}

@article{feng2025learngene,
  title   = {Learngene: Inheritable ``genes'' in intelligent agents},
  author  = {Feng, Fu and Wang, Jing and Yang, Xu and Geng, Xin},
  journal = {Artificial Intelligence},
  volume  = {348},
  pages   = {104421},
  year    = {2025},
  doi     = {10.1016/j.artint.2025.104421}
}

@article{assumpcao2025codeevolve,
  title = {{CodeEvolve}: An Open Source Evolutionary Coding Agent for Algorithmic Discovery and Optimization},
  author = {Henrique Assump{\c{c}}{\~a}o and Diego Ferreira and Leandro Campos and Fabricio Murai},
  journal = {arXiv preprint arXiv:2510.14150},
  year = {2025}
}

@article{wan2025loongflow,
  title = {{LoongFlow}: Directed Evolutionary Search via a Cognitive Plan-Execute-Summarize Paradigm},
  author = {Chunhui Wan and Xunan Dai and Zhuo Wang and Minglei Li and Yanpeng Wang and Yinan Mao and Yu Lan and Zhiwen Xiao},
  journal = {arXiv preprint arXiv:2512.24077},
  year = {2025}
}

@article{foster2026rpm,
  title = {{AI} Research Preference Models},
  author = {Foster, Thomas Simon and Al Omari, Bassel and Fu, Tingchen and Mann, Thomas and Domond, Carl and Cipolina-Kun, Lucia and Gauri, Bhavul and Aghamelu, Muna and Goldie, Alexander D. and Helenowski, Eryk and Gagnon-Audet, Jean-Christophe and Pepe, Alberto and Nazir, Saba and Izcovich, Daniel and Levi, Noam and Hazra, Rishi and Hambardzumyan, Karen and Baldwin, Nicolas and Li, Xian and Josifoski, Martin and Giampouras, Paris and Jalili Sabet, Masoud and Sims, Anya and Momand, Hela and Shavrina, Tatiana and Magka, Despoina and Weston, Jason and Wang, Yulin and Goyal, Anirudh and Henriques, Jo{\~a}o and Bachrach, Yoram and McMilin, Emily and Foerster, Jakob Nicolaus},
  journal = {arXiv preprint arXiv:2608.13940},
  year = {2026}
}

@article{hutter2009paramils,
  title = {{ParamILS}: An Automatic Algorithm Configuration Framework},
  author = {Hutter, Frank and Hoos, Holger H. and Leyton-Brown, Kevin and St{\"u}tzle, Thomas},
  journal = {Journal of Artificial Intelligence Research},
  volume = {36},
  pages = {267--306},
  year = {2009}
}

@article{lopez2016irace,
  title = {The {irace} package: Iterated Racing for Automatic Algorithm Configuration},
  author = {L{\'o}pez-Ib{\'a}{\~n}ez, Manuel and Dubois-Lacoste, J{\'e}r{\'e}mie and P{\'e}rez C{\'a}ceres, Leslie and St{\"u}tzle, Thomas and Birattari, Mauro},
  journal = {Operations Research Perspectives},
  volume = {3},
  pages = {43--58},
  year = {2016},
  doi = {10.1016/j.orp.2016.09.002}
}

@inproceedings{Gagliolo:04ecml,
author="Gagliolo, Matteo
and Zhumatiy, Viktor
and Schmidhuber, J{\"u}rgen",
title="Adaptive Online Time Allocation to Search Algorithms",
booktitle="Machine Learning: ECML 2004",
year="2004",
publisher="Springer Berlin Heidelberg",
address="Berlin, Heidelberg",
pages="134--143",
}

@inproceedings{Gagliolo:05icann,
author = {Gagliolo, Matteo and Schmidhuber, J{\"u}rgen},
title = {A Neural Network Model for Inter-problem Adaptive Online Time Allocation},
booktitle = {Artificial Neural Networks:
Formal Models and Their Applications -- ICANN 2005},
publisher={Springer Berlin Heidelberg},
pages={7--12},
year = {2005}}

@inproceedings{DBLP:conf/dcai/GaglioloS08,
  author    = {Matteo Gagliolo and J{\"u}rgen Schmidhuber},
  title     = {Towards Distributed Algorithm Portfolios},
  booktitle = {International Symposium on Distributed Computing and
               Artificial Intelligence 2008 ({DCAI} 2008)},
  volume    = {50},
  series = {Advances in Soft Computing},
  pages     = {634--643},
  publisher = {Springer},
  address   = {Berlin, Heidelberg},
  year      = {2009},
  doi       = {10.1007/978-3-540-85863-8_75}
}
